\documentclass[runningheads]{llncs}

\usepackage{eccv}

\usepackage{eccvabbrv}

\usepackage{graphicx}
\usepackage{booktabs}
\usepackage{multirow}
\usepackage{makecell}
\usepackage{float}
\usepackage{marvosym}
\usepackage[accsupp]{axessibility}  

\usepackage{hyperref}

\usepackage{orcidlink}

\begin{document}

\title{LMT-GP: Combined Latent Mean-Teacher and Gaussian Process for Semi-supervised Low-light Image Enhancement} 

\titlerunning{Combined Latent Mean-Teacher and GP for Semi-supervised LLIE}

\author{Ye Yu\inst{1}\orcidlink{0000-0002-8646-8426} \and Fengxin Chen\inst{1}\orcidlink{0009-0008-4678-3051} \and
Jun Yu\inst{2}\textsuperscript{(\Letter )}\orcidlink{0000-0002-3197-8103} \and
Zhen Kan\inst{2}\orcidlink{0000-0003-2069-9544}
}

\authorrunning{Yu et al.}

\institute{School of Computer and Information, Hefei University of Technology, Hefei, China
\email{yuye@hfut.edu.cn, fengxin.chen@mail.hfut.edu.cn}\and
Department of Automation, University of Science and Technology of China, Hefei, China\\
\email{\{harryjun,zkan\}@ustc.edu.cn}\\
}
\maketitle

\begin{abstract}
While recent low-light image enhancement (LLIE) methods have made significant advancements, they still face challenges in terms of low visual quality and weak generalization ability when applied to complex scenarios. To address these issues, we propose a semi-supervised method based on latent mean-teacher and Gaussian process, named LMT-GP. We first design a latent mean-teacher framework that integrates both labeled and unlabeled data, as well as their latent vectors, into model training. Meanwhile, we use a mean-teacher-assisted Gaussian process learning strategy to establish a connection between the latent and pseudo-latent vectors obtained from the labeled and unlabeled data. To guide the learning process, we utilize an assisted Gaussian process regression (GPR) loss function.  Furthermore, we design a pseudo-label adaptation module (PAM) to ensure the reliability of the network learning. To demonstrate our method's generalization ability and effectiveness, we apply it to multiple LLIE datasets and high-level vision tasks. Experiment results demonstrate that our method achieves high generalization performance and image quality. The code is available at \url{https://github.com/HFUT-CV/LMT-GP}
\end{abstract}

\section{Introduction}

\begin{figure}[t]
 \centering
    \begin{subfigure}{0.64\linewidth}
    \includegraphics[width=\linewidth]{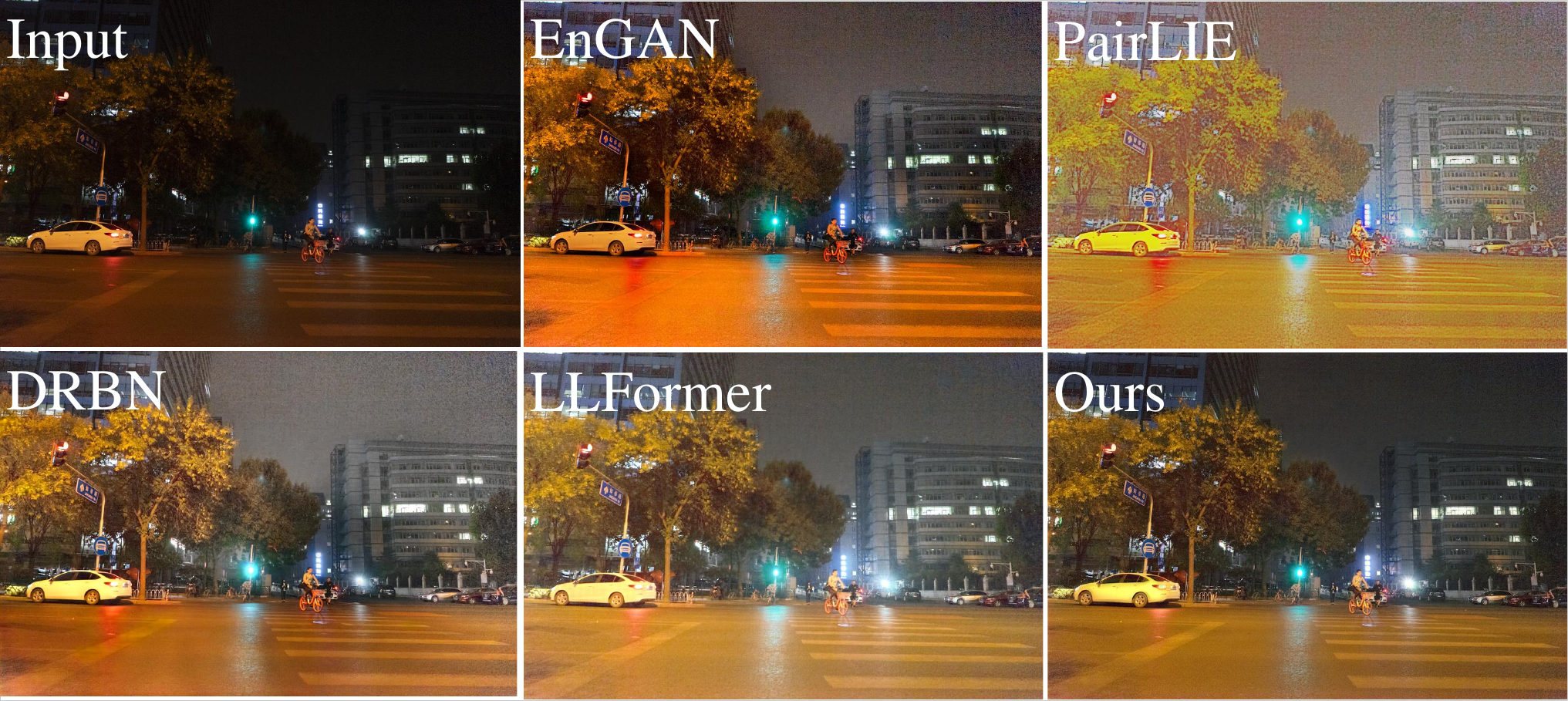}
    \caption{}
    \end{subfigure}
         \hfill
    \begin{subfigure}{0.33\linewidth}
    \includegraphics[width=\linewidth]{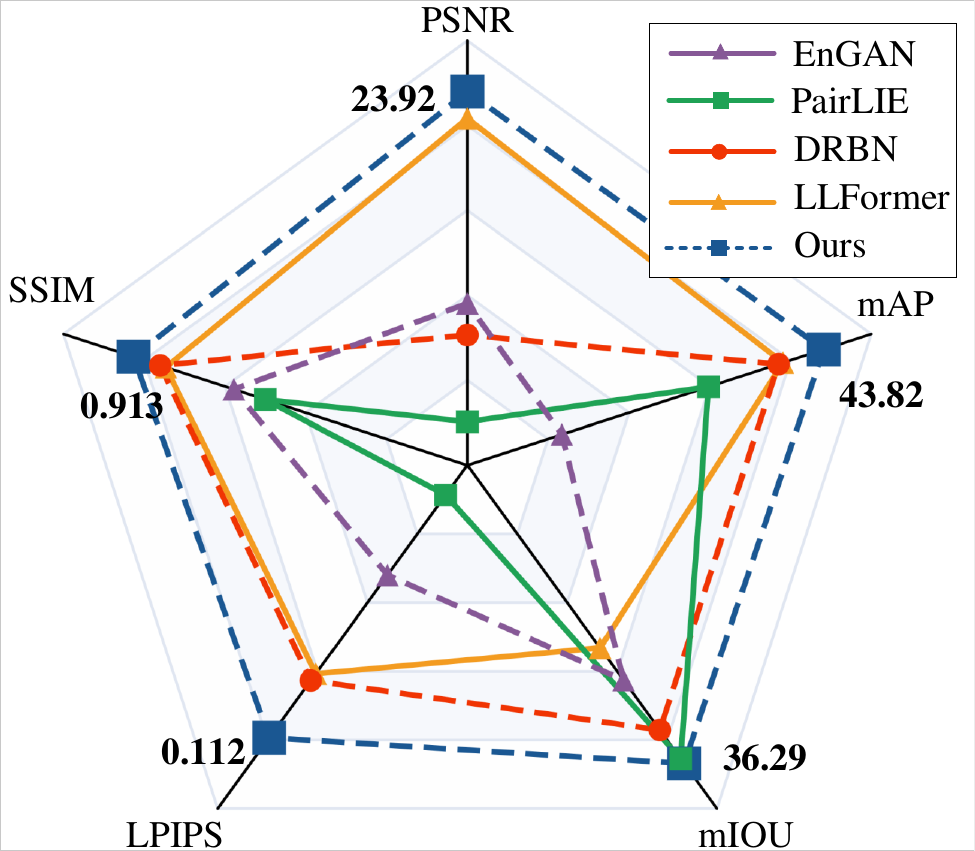}
    \caption{}
    \end{subfigure}
\caption{(a) A comparison on the real-world night images shows that existing LLIE methods have issues such as under/over-enhanced predictions and noisy ouputs. (b) Performance of the compared methods on three tasks: enhancement (measured by PSNR, SSIM, and LPIPS), detection (measured by mAP), and segmentation (measured by mIoU). }
\label{fig:firstfig}
\vspace{-1cm}
\end{figure}

Low-light images are often captured under imperfect lighting conditions, such as backlighting, underexposure, and low-light environments. This can result in poor image quality, including low visibility, low contrast, and excessive noise. Such poor quality images introduce bad user experience and make it challenging for machine vision analysis. For example, they can generate inaccurate results on subsequent image analysis tasks such as object detection and segmentation. To tackle these issues and improve the image quality of low-light images, it is essential to develop low-light image enhancement (LLIE) methods.\cite{bib1, bib2}.

Traditional methods of LLIE\cite{bib3,bib4,bib5,bib6,bib9,bib11} rely on optimization rules and  manual priors. However, these methods are susceptible to producing poor image quality under different lighting conditions, which lead to a loss of details and distorted colors. Supervised deep learning-based LLIE methods\cite{MBLLEN, RetinexNet, GLADNet, DeepUPE, KinD, LLFlow, LLFormer} generally use labeled data to improve image quality. However, capturing these labeled data is challenging and they usually have overfitting problems and limited generalization ability. Some methods\cite{EnlightenGAN, PairLIE, Zero-DCE, Zero-DCE++, RUAS, RetinexDIP, SCI, DRBN} do not use labeled data, but they often have issues such as under/over-enhanced predictions, noisy outputs, and obvious artifacts. Thus, we aim to bridge this gap by using labeled data and unlabeled data for network learning.

The mean-teacher framework\cite{MeanTeacher, mt-detection1, mt-detection2, mt-caption, mt-disloss} is commonly used in semi-supervised learning as it can effectively utilize labeled and unlabeled data for network learning. However, it is difficult to learn the low-level features of unlabeled data  since the mean-teacher only utilize pseudo-labels to supervise the network learning. Furthermore, it has domain shift issues when applied to complex domains. Unfortunately, the degradation of low-light images in real-world is diverse which lead to complex domains\cite{domainissue}. Thus,  we can not directly apply the method to LLIE tasks.

To address these issues, we propose a combined Latent Mean-Teacher and Gaussian Process method (LMT-GP) for semi-supervised low-light image enhancement. We introduce a latent mean-teacher framework as shown in \cref{fig:book} to learn the low-level features of unlabeled data. It incorporates latent vectors of labeled and unlabeled data to supervise the network learning. Meanwhile, we design a mean-teacher-assisted GP learning strategy to avoid the domain shift issue. This strategy leverages labeled data to represent the distribution of latent vectors in unlabeled data. To supervise the mean-teacher-assisted GP learning, we also design an assisted Gaussian process regression (GPR) loss function . Furthermore, we design a pseudo-label adaptation module (PAM) to pick the latent vectors that can guarantee the reliability of the network learning for unlabeled data. Our contributions can be summarized as follows:

\begin{figure}[t]
 \centering
    \begin{subfigure}{0.392\linewidth}
    \includegraphics[width=\linewidth]{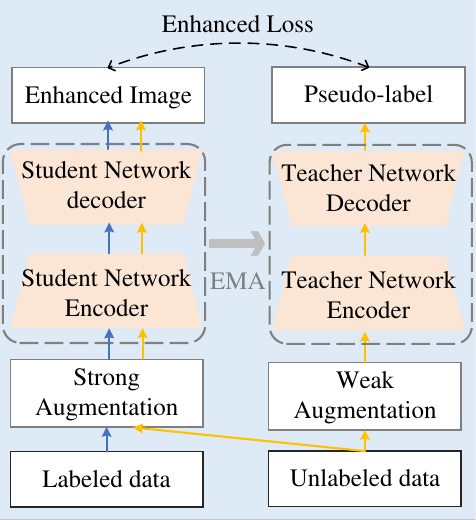}
    \caption{}
    \label{fig:bookfig1}
    \end{subfigure}
         \hfill
    \begin{subfigure}{0.59\linewidth}
    \includegraphics[width=\linewidth]{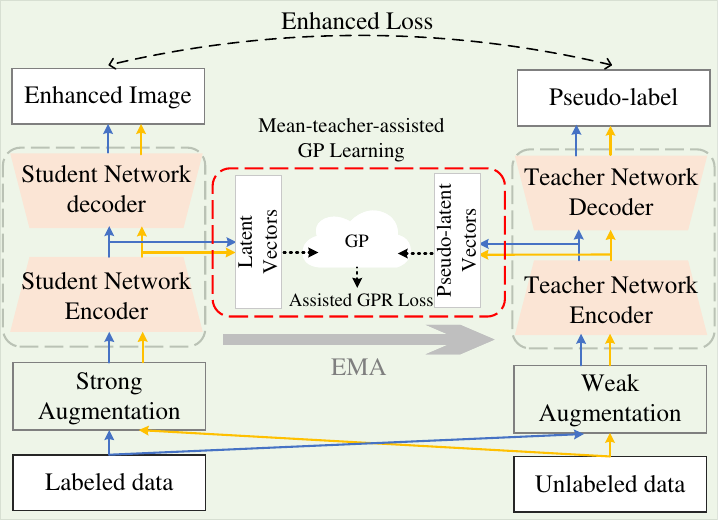}
    \caption{}
    \label{fig:bookfig2}
    \end{subfigure}
\caption{(a) Basic mean-teacher framework. (b) Latent mean-teacher framework. Compared with basic mean-teacher framework, there are three differences: 1) the labeled data needs to be input into both the teacher and student networks; 2) the latent vectors and pseudo-latent vectors are used for network learning; 3) and the mean-teacher-assisted GP learning is performed on latent vectors and pseudo-latent vectors.}
\label{fig:book}
\vspace{-0.5cm}
\end{figure}

\begin{itemize} 
\item {We propose a latent mean-teacher framework to make effective use of labeled and unlabeled data by incorporating latent vectors. We believe this framework can contribute to the development of LLIE methods.}
\item {We design a mean-teacher-assisted GP learning strategy to address the domain shift issue and a assisted GPR loss to guide the learning process. According to our knowledge, this is the first time that mean-teacher and GP have been integrated and applied in the domain of LLIE methods.}
\item {We design a PAM to select latent vectors for the reliability of the network learning. It is proved to be a reliable technique to select supervised information for unlabeled data.}
\item {We conduct experiments on LLIE datasets and high-level vision tasks to show the generalization ability of our method. Results demonstrate that our method effectively enhances the image quality and the generalization ability.}
\end{itemize}

\section{Related Work}

\textbf{Supervised LLIE. }Supervised LLIE methods primarily focus on refining images by minimizing the pixel-wise loss between enhanced images and ground truth (GT) images\cite{SMGLLIE}. For instance, some methods\cite{RetinexNet, DeepUPE, LLNet} applied the Retinex theory to deep neural networks and combined enhanced images with GT images to construct the loss function. Other methods\cite{MBLLEN, GLADNet, LLFormer} chose to directly train end-to-end networks and learn the mapping of low-light images to enhanced images. Compared with minimizing the pixel-wise loss, Wang et al.\cite{LLFlow} proposed a normalizing flow method that effectively imposes constraints on the enhanced image manifold by minimizing negative log-likelihood. Although these methods improve the accuracy on some datasets, they can not achieve good performance in unknown scenarios since they are trained on known paired image datasets.

\noindent\textbf{Unsupervised LLIE. }Unsupervised methods investigate non-paired images rather than low-light images and their corresponding regular-light images to construct networks. Jiang et al.\cite{EnlightenGAN} built a generative adversarial network (GAN) for LLIE using low-light images and non-paired normal-light images. Fu et al.\cite{PairLIE} utilized prior knowledge of low-light images and other low-light images with only different exposure levels to design a self-supervised mechanism. Despite the improved generalization performance, these methods fail to capture the inherent structures of image signals.

\noindent\textbf{Zero-shot LLIE. } Many studies explore the implementation of LLIE by 
solely utilizing low-light images and carefully designing self-learning functions. Guo et al.\cite{Zero-DCE} achieved illumination enhancement through image-specific curve estimation learning. Furthermore, many works studied 
the performance of different architectures, including deep image priors\cite{RetinexDIP}, self-calibrating networks\cite{SCI}, reward-penalty mechanism\cite{xiaomi}, and collaborative search structures\cite{RUAS}. Due to the requirement of meticulously crafted loss functions, the learning process for these methods often becomes unstable.

\noindent\textbf{Semi-supervised LLIE. }Semi-supervised methods have been widely applied to image deraining\cite{semiderain1, semiderain2, Syn2real}, image dehazing\cite{semidehaze1,semidehaze2,semidehaze3}, and image restoration\cite{semires1, SemiUIR}. However, semi-supervised LLIE methods are still in the early stages of research. Yang et.al\cite{DRBN} used a semi-supervised learning method that     combines useful knowledge from paired and unpaired datasets to provide perceptual guidance for detailed signal modeling. However, due to the use of adversarial learning in the second stage, artifacts or noise can be introduced in complex lighting scenarios.

\section{Method}
\subsection{Preliminaries of GP and GPR}
GP\cite{Syn2real} refers to a collection of random variables, where any finite subset of these random variables follows a joint Gaussian distribution. A GP is specified by its mean function $\mu(V)$ and covariance function $K(V, V')$. It can be denoted as follows:
\begin{equation}
\label{eq1}
  \vspace{-0.1cm}
    \{f(V_i)\} \sim N(\mu(V), K(V, V')).
\end{equation}

Let $\Omega=\left\{\left(x_i, y_i\right)\right\}_{i=1}^n$ denotes the training set. The input values $\mathcal{X}=[x_1, x_2, $ $
\ldots, x_n]^{\mathrm{T}}$ and the output values $\mathcal{Y}=\left[y_1, y_2, \ldots, y_n\right]^{\mathrm{T}}$,  where $\mathcal{X}, \mathcal{Y} \in \mathbb{R}^{n \times p}$. Note that the training data are distributed independently and identically. Suppose there are two constraints: 1) the functional relationship between $x_i$ and $y_i$ is given by the function $f(\cdot)$ and is perturbed by Gaussian noise; 2) $\{f(x_i)\}$ is a GP and $y_i$ is defined as follows:
\vspace{-0.3cm}
\begin{equation}
\label{eq2}
    y_i=f\left(x_i\right)+\varepsilon_i.
    \vspace{-0.2cm}
\end{equation}

Based on the training set $\Omega$ and the constraints, we can obtain a joint probability distribution of the test points $\mathcal{X}^*=\left[x_1^*, x_2^*, \ldots, x_m^*\right]^T$ , the unknown function values $f^*=\left\lceil f\left(x_1^*\right), f\left(x_2^*\right), \ldots, f\left(x_m^*\right)\right]^T$ and $\mathcal{Y}$. This joint probability distribution is defined as follows:
\begin{equation}
\vspace{-0.2cm}
\label{eq3}
 \left[\begin{array}{c}
\mathcal{Y} \\
f^* 
\end{array} \right] \sim N\left(\left[\begin{array}{c}
\mu\left(\mathcal{X}\right) \\
\mu\left(\mathcal{X}^*\right)
\end{array}\right], 
   \begin{array}{c}
        \left[\begin{array}{cc}
K(\mathcal{X}, \mathcal{X})+\sigma^2 \boldsymbol{I} & K(\mathcal{X}^*, \mathcal{X}) \\
K\left(\mathcal{X}^*, \mathcal{X}\right) & K\left(\mathcal{X}^*, \mathcal{X}^*\right)+\sigma^2 \boldsymbol{I}
\end{array}
\right]
    \end{array}
\right), 
\end{equation}
where $\boldsymbol{I}$ is the identity matrix and $\mu\left(\cdot\right)\text{=}0$ in this case. 

According to the properties of joint Gaussian distribution, the conditional probability density of $f^*$ is defined as follows:
\vspace{-0.2cm}
\begin{equation}
    p\left(f^* \mid \mathcal{X}^*, \mathcal{X}, \mathcal{Y}\right) \sim N\left(\mu^*, \Sigma^*\right),
    \vspace{-0.1cm}
    \label{eq4}
\end{equation}
where 
\vspace{-0.1cm}
\begin{equation}
\mu^*=K\left(\mathcal{X}^*, \mathcal{X}\right)\left(K(\mathcal{X}, \mathcal{X})+\sigma^2 I\right)^{-1} \mathcal{Y},
\vspace{-0.2cm}
\label{eq5}
\end{equation}
\begin{equation}
    \Sigma^*= K(\mathcal{X}^*, \mathcal{X}^*)-K\left(\mathcal{X}^*, \mathcal{X}\right)\left(K(\mathcal{X}, \mathcal{X}) +\sigma^2 I\right)^{-1} 
    K(\mathcal{X}, \mathcal{X}^*)+\sigma^2 I.
    \vspace{-0.3cm}
    \label{eq63}
\end{equation}

\subsection{Problem formulation}
Let $\Omega_{\mathrm{L}}=\left\{\left(x_l^i, y_l^i\right) \mid x_l^i \in \mathrm{X}_{\mathrm{L}}, \mathrm{y}_l^i \in \mathrm{Y}_{\mathrm{L}}\right\}_{i=1}^N$ denotes the labeled dataset, where $\mathrm{X}_{\mathrm{L}}$ and $\mathrm{Y}_{\mathrm{L}}$ are sets of low-light images and normal-light images, respectively; $x_l^i$ and $y_l^i$ are paired data. Similarly, $\Omega_{\mathrm{U}}=\left\{x_{u}^i \mid x_u^i \in \mathrm{X}_{\mathrm{U}}\right\}_{i=1}^N$ represents the unlabeled dataset consisting of independent low-light images. Note that $\Omega_{\mathrm{L}} \cap \Omega_{\mathrm{U}}=\varnothing$. Our goal is to establish a mapping relationship on the joint dataset $\Omega=\Omega_{\mathrm{L}}\cup \Omega_{\mathrm{U}}$ that facilitates the conversion of low-light images into normal-light images and fully utilizes the latent information contained in unlabeled data.

\label{sec:method}
\begin{figure}[t]
    \centering
    \includegraphics[width=\linewidth]{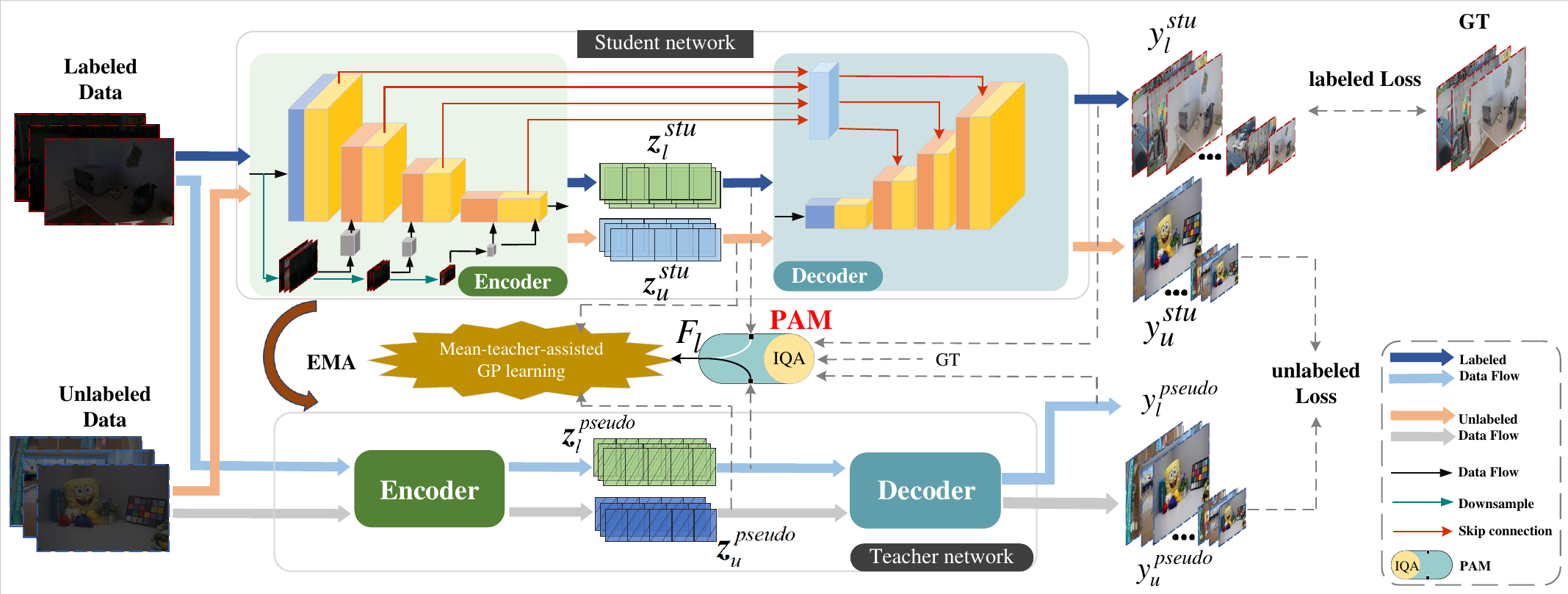}
    \caption{Overview of LMT-GP. LMT-GP is based on the latent mean-teacher framework. Labeled and unlabeled data are simultaneously input into the student network and the teacher network. The student network generates latent vectors $z_l^{stu}$ and $z_u^{stu}$ for labeled and unlabeled data, respectively. The teacher network generates pseudo-latent vectors $z_l^{pseudo}$ and $z_u^{pseudo}$ for labeled and unlabeled data, respectively. After being processed by PAM, we perform mean-teacher-assisted GP learning on the latent vectors, pseudo-latent vectors and the generated vector sequence $F_l$.}
    \label{mainView}
    \vspace{-0.4cm}
\end{figure}
  \vspace{-0.3cm}
\subsection{LMT-GP}
\cref{mainView} shows the LMT-GP. It consists of a teacher network, a student network, and PAM. The teacher and student network have the identical architecture, which consists of an encoder and a decoder. The encoder generates intermediate features and latent vectors. The decoder then utilizes these intermediate features and latent vectors to generate multiscale output images. More details about the architecture can be found in the Supplementary Materials.

To help the unlabeled data learn low-level features, our framework incorporates labeled, unlabeled data, and the corresponding latent vectors into the network training. Labeled data and unlabeled data are input to the student and teacher networks simultaneously during the training process. They follow different data pipelines as illustrated in \cref{mainView}. For labeled images $x_l \in \Omega_{\mathrm{L}}$, the teacher network generates $K$-scale pseudo-labels $y_l^{pseudo}=\left\{y_{l,j}^{pseudo}\right\}_{j=1}^K$ (where $K=4$ in this paper) and pseudo-latent vectors $z_l^{pseudo}$. Meanwhile, the student network generates the corresponding $K$-scale output images $y_l^{stu}=\left\{y_{l,j}^{stu}\right\}_{j=1}^K$ and latent vectors $z_l^{stu}$. Similarly, for unlabeled images $x_u \in \Omega_{\mathrm{U}}$, the teacher network generates $y_u^{pseudo}$ and $z_u^{pseudo}$. Meanwhile, the student network generates $y_u^{stu}$ and $z_u^{stu}$. Then $z_l^{pseudo}$, $z_l^{stu}$, $y_l^{stu}$, and $y_l^{pseudo}$ are passed through the PAM to generate vector sequence $F_l=\{z^i_r\}_{i=1}^N$. Subsequently, we perform mean-teacher-assisted GP learning on $F_l$, $z_u^{pseudo}$, and $z_u^{stu}$. Furthermore, we design a labeled loss between $y_l^{stu}$ and GT images. We also design an unlabeled loss between $y_u^{stu}$ and $y_u^{pseudo}$.

Note that the weight $\boldsymbol{\theta}_{\boldsymbol{s}}$ of the student network is updated through the network learning. After each epoch, the weight $\boldsymbol{\theta}_{\boldsymbol{t}}$ of the teacher network is updated using the exponential moving average\cite{SemiUIR} (EMA) of $\boldsymbol{\theta}_{\boldsymbol{s}}$ as follows:
\begin{equation}
    \boldsymbol{\theta}_{\boldsymbol{t}}=\eta \boldsymbol{\theta}_{\boldsymbol{t}}+(\mathbf{1}-\eta) \boldsymbol{\theta}_{\boldsymbol{s}},
\end{equation}
where $\eta \in(\mathbf{0 , 1})$ is the momentum.
\vspace{-0.1cm}
\subsection{Mean-Teacher-Assisted GP Learning} 
To address the domain shift issue of mean-teacher and improve generalization ability, we utilize mean-teacher-assisted GP learning to model labeled and unlabeled data. We first employ principal component analysis (PCA) to select the most representative $n$ vectors from $F_{l}$. Then we use them to form $\hat{F}_l=\{\hat{\mathit{z}}_r^i,...,\hat{\mathit{z}}_r^n\}$, working as a set of basis vectors. The latent vectors of unlabeled data is a linear combination of vectors in $\hat{F}_l$ and noise interference $\varepsilon^m$. It can be represented as
\vspace{-0.1cm}
\begin{equation}
    z_u^{stu,m}=\sum_{i=1}^{N} \alpha_i^m \hat{\mathit{z}}_r^i+\varepsilon^m,
    \vspace{-0.2cm}
\end{equation}
where $\varepsilon^m \sim N\left(0, \sigma^2\right)$, $\alpha_i^m$ is the coefficient and $z_u^{stu,m}$ represents the latent vector obtained by inputting the $m$-th image from the unlabeled data into the student network.

We assume the distribution of $\hat{F}_l$ follows a GP, which implies that the distribution of $\{\hat{\mathit{z}}_r^i,...,\hat{\mathit{z}}_r^n\} \cup\left\{z_u^j\right\}_{j=1}^M$ also follows a GP. We use the GPR to jointly model the distribution of latent vectors for both labeled and unlabeled data as
\begin{equation}
\left[\begin{array}{l}
\hat{\boldsymbol{z}}_r \\
\boldsymbol{z}_u
\end{array}\right] = N\left(\left[\begin{array}{l}
\hat{\boldsymbol{\mu}}_r \\
\boldsymbol{\mu}_u
\end{array}\right],
\left[\begin{array}{cc}
K\left(\hat{\boldsymbol{z}}_r, \hat{\boldsymbol{z}}_r\right)+\sigma_{\varepsilon}^2 \boldsymbol{I} & K\left(\hat{\boldsymbol{z}}_r, \boldsymbol{z}_u\right) \\
K\left(\boldsymbol{z}_u, \hat{\boldsymbol{z}}_r\right) & K\left(\boldsymbol{z}_u, \boldsymbol{z}_u\right)+\sigma_{\varepsilon}^2 \boldsymbol{I}
\end{array}\right]\right),
\vspace{-0.1cm}
\end{equation}
where $K(\cdot, \cdot)$ is a kernel function that is defined as follows:
\vspace{-0.1cm}
\begin{equation}
\vspace{-0.2cm}
    K\left(z, z^{\prime}\right)=\exp \left(-\frac{\left(z-z^{\prime}\right)^2}{2}\right).
\end{equation}
Using \cref{eq4}-\cref{eq63}, the posterior distribution of the unlabeled latent vectors can be represented as
\vspace{-0.1cm}
\begin{equation}
    p\left(z_u^m \mid \Omega_{\mathrm{L}}, \hat{F}_l\right) \sim N\left(\mu_u^m, \sum{}_u^m\right),
\end{equation}
where
\vspace{-0.1cm}
\begin{equation}
    \mu_u^m=K\left(z_u^m, \hat{F}_l\right)\left(K\left(\hat{F}_l, \hat{F}_l\right)+\sigma^2 I\right)^{-1} \hat{F}_l,
    \vspace{-0.25cm}
\end{equation}
\begin{equation}
       \sum{}_u^m=K\left(z_u^m, z_u^m\right)-K\left(z_u^m, \hat{F}_l\right)\left(K\left(\hat{F}_l, \hat{F}_l\right)  
   +\sigma^2 I\right)^{-1} K\left(\hat{F}_l, z_u^m\right)+\sigma^2 I.
   \vspace{-0.1cm}
\end{equation}

We maximize the posterior distribution of latent vectors $z_u^m$ of unlabeled data with the pseudo-latent vectors to constrain the learning. Hence, the assisted GPR loss $L_{g pr}$ is defined as follows:
\begin{equation}
    L_{g pr}=\left(z_u^{pseudo, m}-\mu_u^m\right)^{\mathrm{T}}\left(\sum{}_u^m\right)^{-1}\left(z_u^{p s e u d o, m}-\mu_u^m\right) +\log \left|\sum{}_u^m\right|,
\end{equation}
where $z_u^{pseudo,m}$ represents the latent vector obtained by inputting the $m$-th image from the unlabeled data into the teacher network.

Due to the computational complexity of $\sum{}^m_u$, the network learning is unstable. Thus, we set $\sum{}^m_u$ to a constant to reduce its impact. The variant of assisted GPR loss is denoted as follows:
\begin{equation}
    \hat{L}_{gpr}=\left\|z_u^{p s e u d o, m}-\hat{\mu}_u^m\right\|_2^2.
\end{equation}

\subsection{PAM}
It is important to acknowledge that the performance of the teacher network may not always surpass that of the student network in practical settings\cite{SemiUIR}. Consequently, pseudo-latent vectors produced by the teacher network may not be an optimal guide for the student network's learning. Toward this end, we introduce PAM, a module dedicated to selecting latent vectors that help guide the student network's learning. We select latent vectors from $y_l^{stu}$ and $y_l^{pseudo}$, based on the PSNR metric scores. The vector sequence $F_l$ can be defined as follows:
\vspace{-0.1cm}
\begin{equation}
\begin{gathered}
     F_l=\{\mathit{z}^i_r|\mathit{z}^i_r=\Phi(y_l^i,y^{pseudo, i}_l,y_l^{stu,i})\}_{i=1}^N, \\ \Phi(\cdot,\cdot,\cdot)=\{\begin{array}{ll}
     \mathit{z}_l^{pseudo, i},&  \Gamma(y_{l}^{i}, y^{stu,i}_{l})<\Gamma(y^{i}_l, y_l^{pseudo,i}) \\
\mathit{z}_l^{stu,i}, & else
\end{array},
\end{gathered}
\end{equation}
where $\Gamma(\cdot, \cdot)$ represents the PSNR score, $y^{stu,i}_l$ and $y_l^{pseudo,i}$ represent the outputs of the $i$-th labeled data by the student and teacher networks, respectively.

\subsection{Loss Function}

\textbf{Labeled Loss. }We get $K$-scale images $\left\{y_{l,j}\right\}_{j=1}^k$ of the same size as the outputs by performing bilinear interpolation on the original GT. The reconstruction loss $L_1$ can be defined as follows:
\vspace{-0.3cm}
\begin{equation} \label{eq6}
    L_{1}=\sum_{j=\mathbf{1}}^K \zeta_i\left\|y_{l,j}-y_{l,j}^{stu}\right\|_1,
    \vspace{-0.3cm}
\end{equation}
where $\zeta_i$ represents the weighting parameter used for the $j\text{-}th$ scale. To further improve the similarity of image structure and high-frequency details between generated images and the GT, we also adopt negative SSIM loss\cite{SSIM, nSSIM} and perceptual loss\cite{perceptual}, which are defined as follows:
    \vspace{-0.3cm}
\begin{equation}
    L_{s s i m}=K-\sum_{j=\mathbf{1}}^K \zeta_j \operatorname{SSIM}\left(y_{l,j}, y_{l,j}^{stu}\right),
\end{equation}
    \vspace{-0.3cm}
\begin{equation}
    L_p=\sum_{j=1}^K \zeta_j\left\|\Theta\left(y_{l,j}\right)-\Theta\left(y_{l,j}^{stu}\right)\right\|_2^2,
\end{equation}
where $\Theta(\cdot)$ represents the pre-trained VGG-16 network. Total labeled loss $L_{su}$ can be expressed as
\begin{equation}
    L_{s u}=L_1+\lambda_1 L_{s s i m}+\lambda_2 L_p,
\end{equation}
where $\lambda_1$ and $\lambda_2$ are weighting parameters.

\noindent\textbf{Unlabeled Loss. } 
The unlabeled loss guides the student network to learn global information about unlabeled data. The unlabeled Loss $L_{un}$ is presented as follows:
\begin{equation}
     L_{u n}=\sum_{j=1}^K \zeta_j\left\|y_{u,j}^{stu}-y_{u,j}^{pseudo}\right\|_1. 
\end{equation}
The overall loss $L_{total }$ is defined as follows:
\begin{equation}
    L_{total }=L_{s u}+\omega_1 L_{u n}+\omega_2 \hat{L}_{g pr}.
\end{equation}
\vspace{-0.8cm}

\section{Experiment}
\subsection{Experimental Setup}

\textbf{Training Details. }We train the LMT-GP using randomly cropped patches of size 256 $\times$ 256. Meanwhile, we utilize the Adobe Photoshop Lightroom software to synthesize unlabeled data. Specifically, we synthesize 5000 low-light images as unlabeled data from normal-light images of the labeled data by setting the exposure parameter to [-5,0], the contrast parameter to [-100,100], and the vibrance parameter to [-100,0]. In the training process, we use the Adam optimizer with parameters  $\beta_1\text{ = }0.9 \text{, }\beta_2\text{ = }0.999$\text{, and} $\epsilon\text{ = }10^{-8}$. The initial learning rate is $10^{-3}$. The number of training epochs is 600. For the weighting parameters, we empirically set them to $n\text{ = }16\text{,}$ $\lambda_1\text{ = }0.4 \text{, }  \lambda_2\text{ = }0.6$\text{, }$\zeta_1=1\text{, }\zeta_2\text{ = }0.8\text{, } \zeta_3\text{ = }0.6 \text{ and } \zeta_4\text{ = }0.4$. Parameters $\omega_1$ and $\omega_2$ are updated at training epoch $t$ using exponential annealing functions\cite{EWF}, and their final states are $\omega_1\text{ = }0.2$ and $\omega_2\text{ = }0.01$. During the testing, we pad the input images symmetrically on both sides to make them multiples of 32$\times$32. After inference, we crop the padded images back to their original sizes. All experiments are conducted on a PC with a single RTX 4090 GPU.

\noindent\textbf{Compared Methods. }We compare the LMT-GP with 15 state-of-the-art LLIE methods. Specifically, they are categorized into five groups: traditional methods (LIME\cite{LIME}), supervised learning methods (MBLLEN\cite{MBLLEN}, RetinexNet, GLADNet\cite{GLADNet}, DeepUPE\cite{DeepUPE}, KinD\cite{KinD}, LLFlow, and LLFormer\cite{LLFormer}), unsupervised learning methods (EnGAN and PairLIE), zero-shot methods (Zero-DCE++\cite{Zero-DCE++}, RUAS\cite{RUAS}, SCI, and RetinexDIP) and semi-supervised learning methods (DRBN).

\noindent\textbf{Datasets and Metrics. }We evaluate the performance of LLIE on two widely used low-light datasets: the LOL dataset\cite{RetinexNet}(comprising 485 pairs of low- and normal-light images for training and 15 pairs of images for testing) and the VE-LOL dataset\cite{VE-LOL} (containing 100 paired images captured in real scenarios for testing). We use three full-reference metrics: PSNR, SSIM\cite{SSIM}, and LPIPS\cite{LPIPS}, to evaluate the quality of enhanced images. Furthermore, to evaluate the performance for subsequent high-level vision tasks, we conduct separate experiments on the Darkface \cite{Darkface} and ACDC\cite{ACDC} datasets. Darkface comprises 6000 nighttime images, with a training–validation–test split ratio of 8:1:1, and the detection accuracy was evaluated using the mAP metric. ACDC comprises 400 images for training and 106 images for testing, and the semantic segmentation results were evaluated using the mIoU metric. More experimental comparisons can be found in the Supplementary Materials.
\vspace{-0.3cm}
\subsection{Comparison with State-of-the-Art Methods}
To validate the effectiveness of our method in generalization performance, we train on the LOL dataset and test on the LOL, VE-LOL, Darkface, and ACDC datasets separately. Note that we select the top 10 methods with the highest PSNR scores on the LOL dataset when testing on the Darkface and ACDC datasets. Then we apply each of these methods separately to preprocess the Darkface and ACDC datasets. We adopt YOLOv5 as our detector for evaluating on the Darkface dataset and HRNet\cite{HRNet} as the baseline to evaluate on the ACDC dataset.

\begin{figure}[H]
    \centering
    \begin{subfigure}{0.15\linewidth}
    \includegraphics[width=\linewidth]{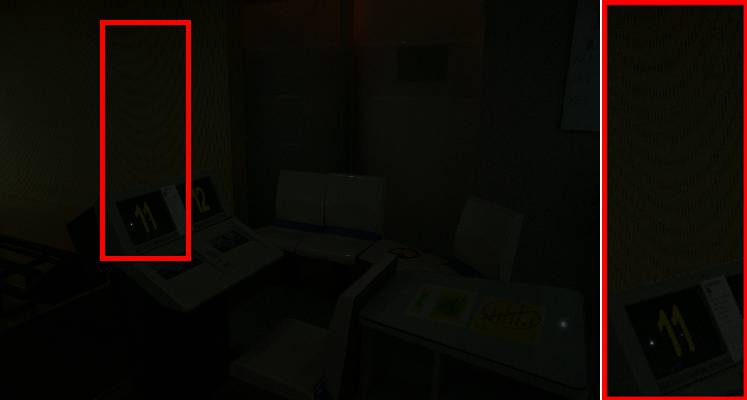}
    \subcaption*{\centering{Input}}
    \label{fig:enter-label1}
    \end{subfigure}
     \hfill
    \begin{subfigure}{0.15\linewidth}
    \includegraphics[width=\linewidth]{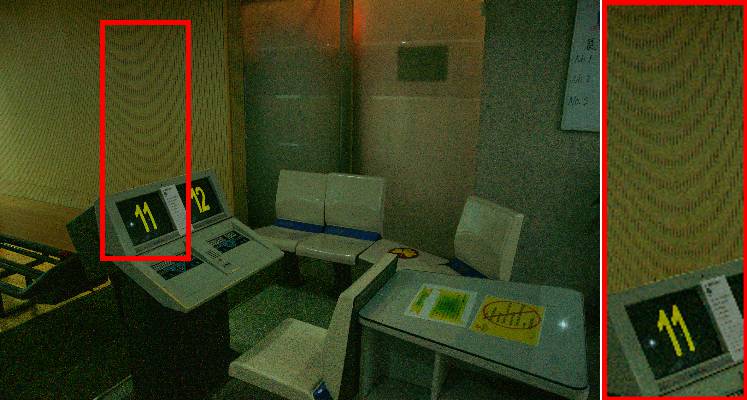}
    \subcaption*{\centering{LIME}}
    \label{fig:enter-label2}
    \end{subfigure}
     \hfill
    \begin{subfigure}{0.15\linewidth}
    \includegraphics[width=\linewidth]{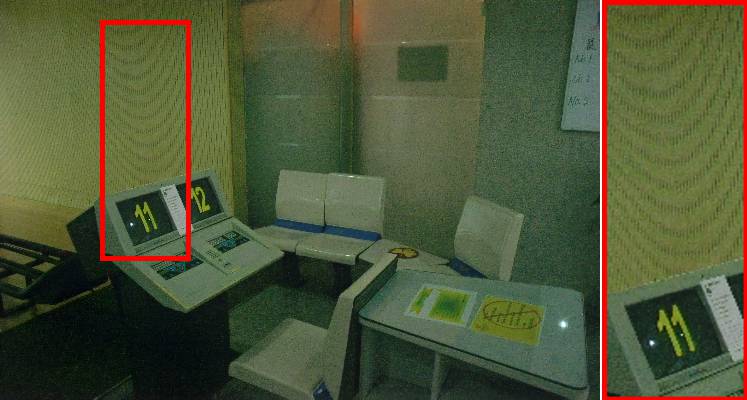}
    \subcaption*{\centering{EnGAN}}
    \label{fig:enter-label3}
    \end{subfigure}
    \hfill
              \begin{subfigure}{0.15\linewidth}
    \includegraphics[width=\linewidth]{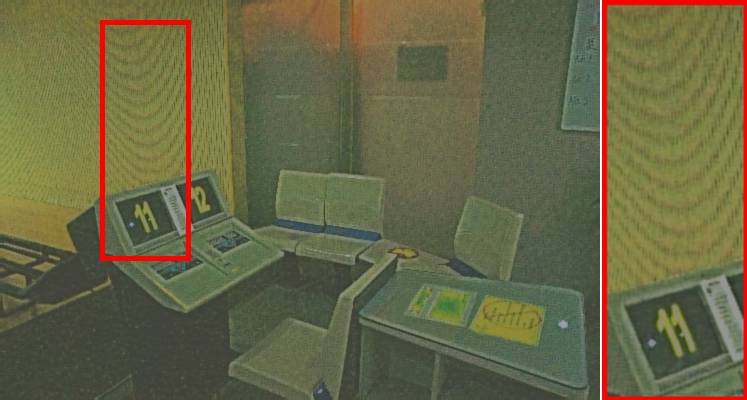}
    \subcaption*{\centering{PairLIE}}
    \label{fig:enter-label5}
    \end{subfigure}
     \hfill
    \begin{subfigure}{0.15\linewidth}
    \includegraphics[width=\linewidth]{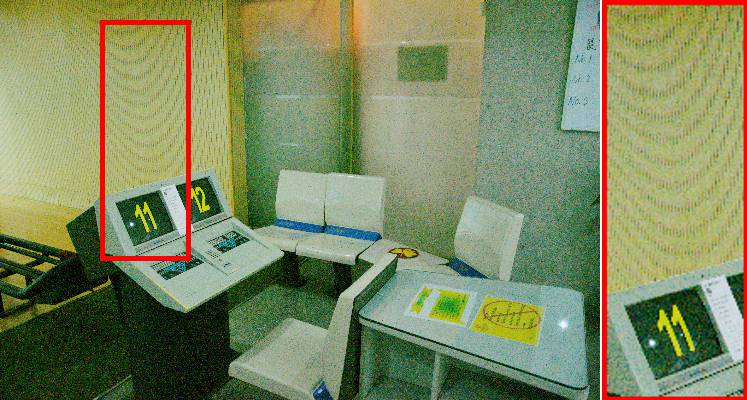}
    \subcaption*{\centering{ZeroDCE++}}
    \label{fig:enter-label6}
    \end{subfigure}
         \hfill
    \begin{subfigure}{0.15\linewidth}
    \includegraphics[width=\linewidth]{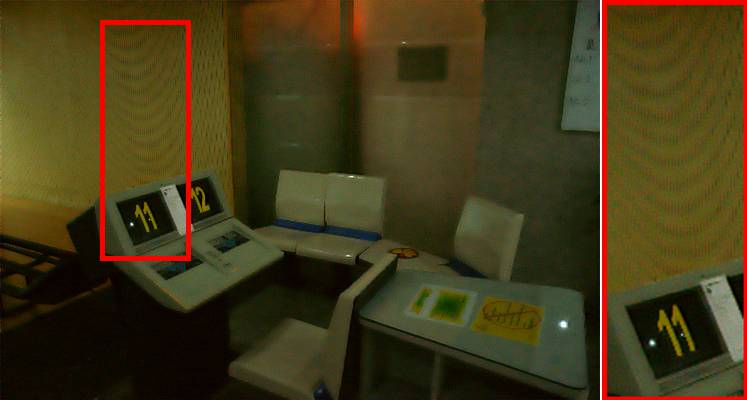}
    \subcaption*{\centering{RUAS}}
    \label{fig:enter-label7}
    \end{subfigure}
         \hfill
                 \begin{subfigure}{0.15\linewidth}
    \includegraphics[width=\linewidth]{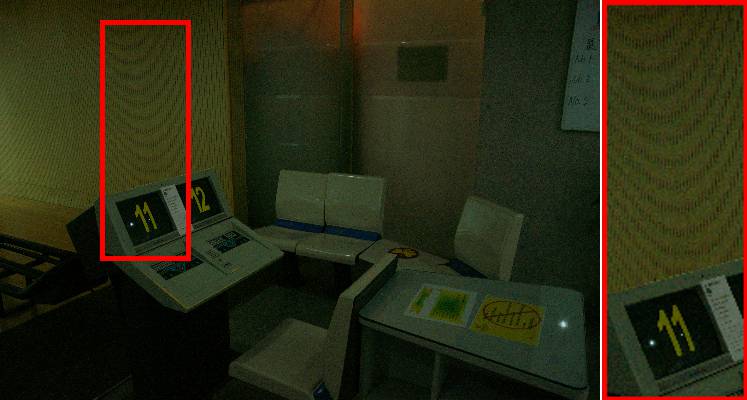}
    \subcaption*{\centering{SCI}}
    \label{fig:enter-label4}
    \end{subfigure}
     \hfill
    \begin{subfigure}{0.15\linewidth}
    \includegraphics[width=\linewidth]{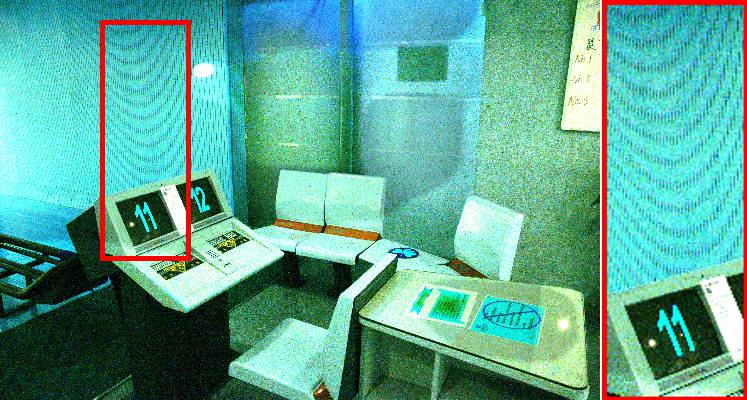}
    \subcaption*{\centering{RetinexDIP}}
    \label{fig:enter-label8}
    \end{subfigure}
         \hfill
    \begin{subfigure}{0.15\linewidth}
    \includegraphics[width=\linewidth]{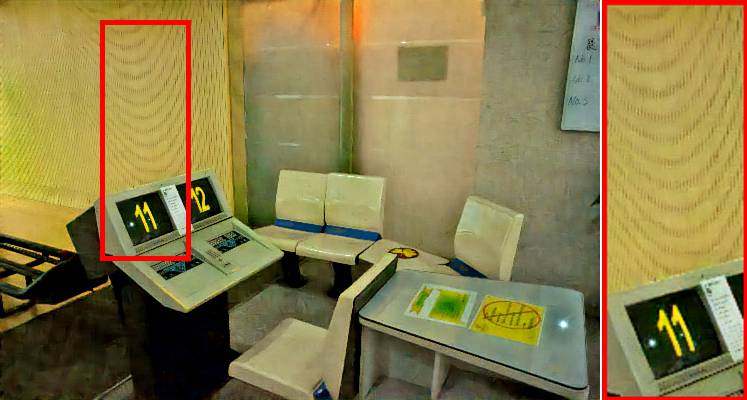}
    \subcaption*{\centering{MBLLEN}}
    \label{fig:enter-label9}
    \end{subfigure}
         \hfill
    \begin{subfigure}{0.15\linewidth}
    \includegraphics[width=\linewidth]{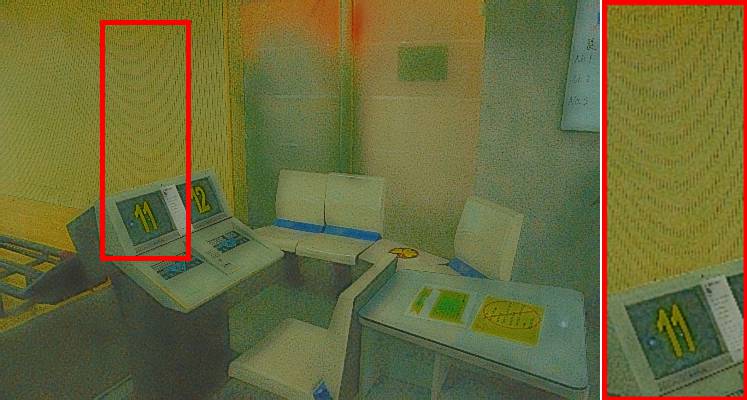}
    \subcaption*{\centering{RetinexNet}}
    \label{fig:enter-label10}
    \end{subfigure}
         \hfill
    \begin{subfigure}{0.15\linewidth}
    \includegraphics[width=\linewidth]{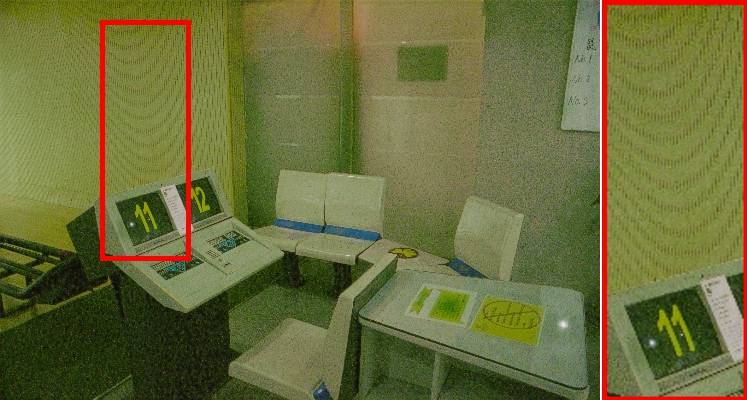}
    \subcaption*{\centering{GLADNet}}
    \label{fig:enter-label11}
    \end{subfigure}
         \hfill
    \begin{subfigure}{0.15\linewidth}
    \includegraphics[width=\linewidth]{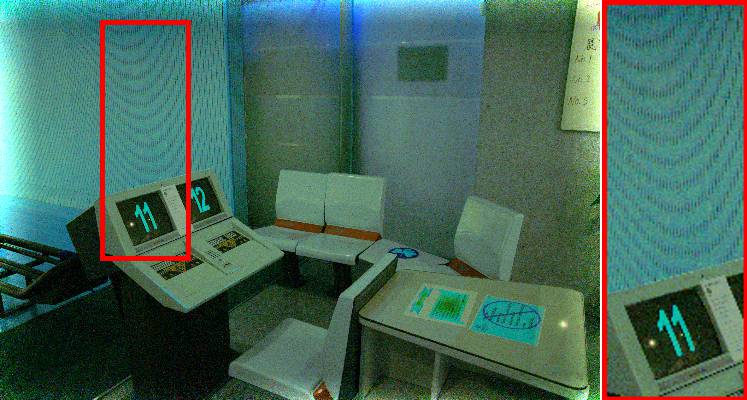}
    \subcaption*{\centering{DeepUPE}}
    \label{fig:enter-label12}
    \end{subfigure}
         \hfill
    \begin{subfigure}{0.15\linewidth}
    \includegraphics[width=\linewidth]{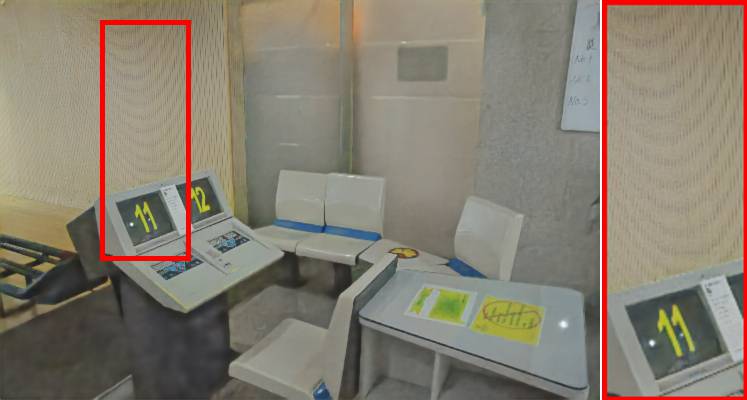}
    \subcaption*{\centering{KinD}}
    \label{fig:enter-label13}
    \end{subfigure}
         \hfill
    \begin{subfigure}{0.15\linewidth}
    \includegraphics[width=\linewidth]{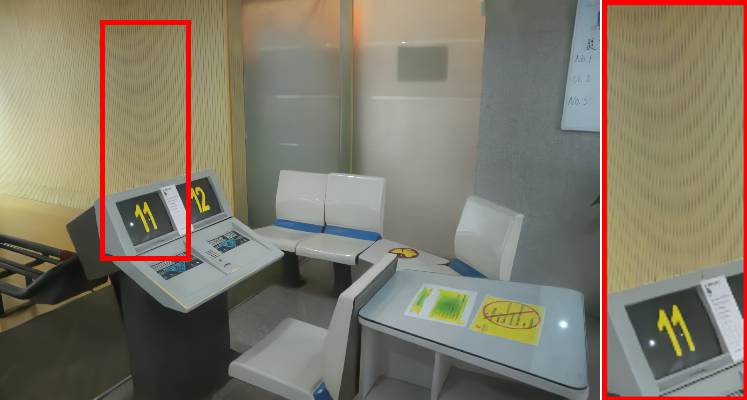}
    \subcaption*{\centering{LLFlow}}
    \label{fig:enter-label14}
    \end{subfigure}
         \hfill
    \begin{subfigure}{0.15\linewidth}
    \includegraphics[width=\linewidth]{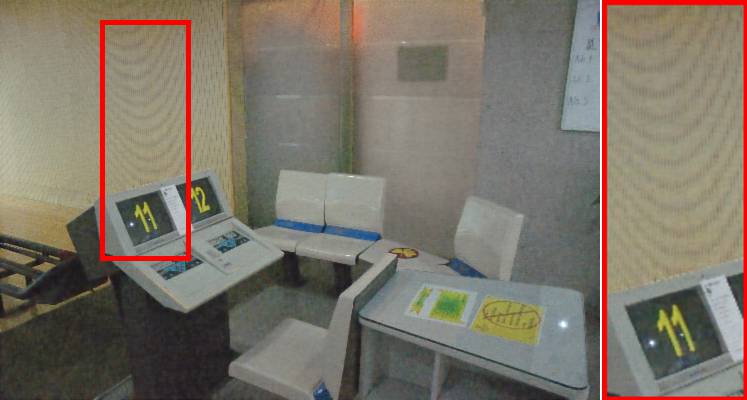}
    \subcaption*{\centering{LLFormer}}
    \label{fig:enter-label15}
    \end{subfigure}
         \hfill
    \begin{subfigure}{0.15\linewidth}
    \includegraphics[width=\linewidth]{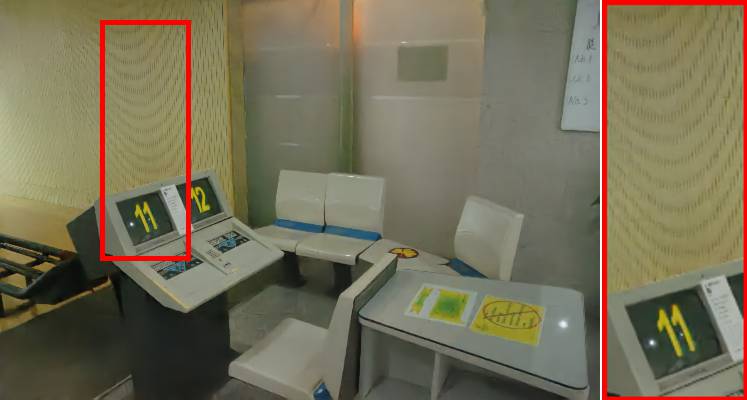}
    \subcaption*{\centering{DRBN}}
    \label{fig:enter-label16}
    \end{subfigure}
         \hfill
    \begin{subfigure}{0.15\linewidth}
    \includegraphics[width=\linewidth]{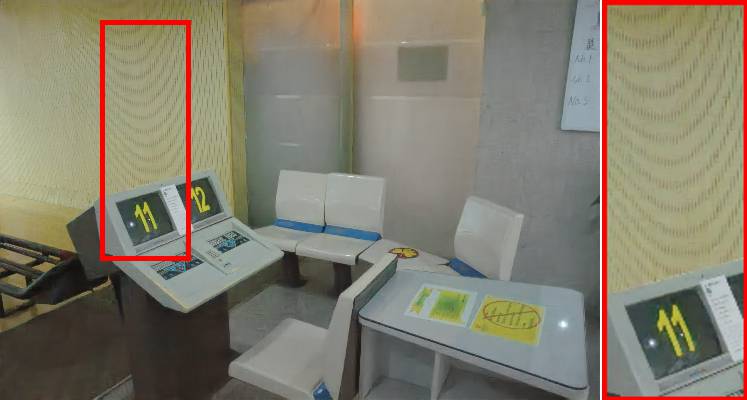}
    \subcaption*{\centering{Ours}}
    \label{fig:enter-label17}
    \end{subfigure}
         \hfill
    \begin{subfigure}{0.15\linewidth}
    \includegraphics[width=\linewidth]{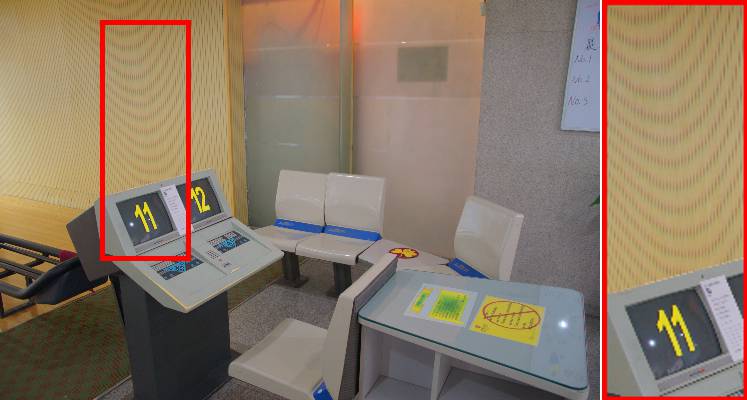}
    \subcaption*{\centering{GT}}
    \label{fig:enter-label18}
    \end{subfigure}
\caption{Visual comparison of state-of-the-art LLIE methods on the LOL dataset. More results can be found in the Supplementary Materials.}
\label{fig:lol}
\end{figure}

\noindent\textbf{Evaluation on LOL. }As shown in \Cref{tab:lol_}, LMT-GP's performance on the LOL dataset is competitive. It yields the highest PSNR and LPIPS scores, surpassing the second-best method by 0.44 dB and 0.018, respectively.  The qualitative results of the test set are shown in \cref{fig:lol}. The images obtained by LIME, 
\begin{table}[H]\scriptsize
 \caption{Quantitative results of enhancement on the LOL and VE-LOL datasets. ``T”, ``U”, ``Z”, ``S”, and ``SE” represent the ``Traditional”, ``Unsupervised”, ``Zero-shot”, ``Supervised”, and ``Semi-supervised” methods, respectively. The best results are in \textcolor{red}{\textbf{red}} whereas the second-best are in \textcolor{blue}{\textbf{blue}}.}
\centering
\renewcommand{\arraystretch}{1.2}
    \begin{tabular}{p{3.4cm}|p{1.1cm}<{\centering}|p{1.15cm}<{\centering}p{1.15cm}<{\centering}p{1.15cm}<{\centering}|p{1.15cm}<{\centering}p{1.15cm}<{\centering}p{1.15cm}<{\centering}}
    \toprule
        \multicolumn{1}{c|}{\multirow{2}{*}{Method}} & \multirow{2}{*}{Type} & \multicolumn{3}{c|}{LOL}  & \multicolumn{3}{c}{VE-LOL}  \\ \cline{3-8}
         &  & PSNR↑ & SSIM↑ & LPIPS↓ & PSNR↑ & SSIM↑ & LPIPS↓ \\ \hline
        LIME\cite{LIME} \scalebox{0.6}{TIP'17} & T & 14.02 & 0.623 & 0.372 & 16.95 & 0.640 & 0.348 \\ \hline
        EnGAN \cite{EnlightenGAN}  \scalebox{0.6}{TIP'21} & U & 19.43 & 0.789 & 0.254 & 18.97 & 0.761 & 0.247 \\ 
        PairLIE \cite{PairLIE}  \scalebox{0.6}{CVPR'23} & U & 16.92 & 0.750 & 0.324 & 18.68 & 0.762 & 0.305 \\ \hline
        ZeroDCE++  \cite{Zero-DCE++}  \scalebox{0.6}{TPAMI'22} & Z & 14.49 & 0.621 & 0.409 & 12.72 & 0.541 & 0.463 \\ 
        RUAS  \cite{RUAS}    \scalebox{0.6}{CVPR'21} & Z & 16.40 & 0.704 & 0.270 & 15.33 & 0.673 & 0.310 \\ 
        SCI  \cite{SCI}   \scalebox{0.6}{CVPR'22} & Z & 13.65 & 0.589 & 0.341 & 16.67 & 0.609 & 0.306 \\ 
        RetinexDIP \cite{RetinexDIP}  \scalebox{0.6}{TCSVT'22} & Z & 10.71 & 0.461 & 0.590 & 11.00 & 0.481 & 0.567 \\ \hline
        MBLLEN    \cite{MBLLEN}  \scalebox{0.6}{BMVC'18} & S & 17.56 & 0.781 & 0.174 & 17.30 & 0.750 & 0.222 \\ 
        RetinexNet   \cite{RetinexNet}  \scalebox{0.6}{BMVC'18} & S & 17.61 & 0.683 & 0.386 & 17.71 & 0.679 & 0.439 \\ 
        GLADNet    \cite{GLADNet}   \scalebox{0.6}{FG'18} & S & 19.72 & 0.731 & 0.321 & 16.49 & 0.735 & 0.478 \\ 
        DeepUPE   \cite{DeepUPE}  \scalebox{0.6}{CVPR'19} & S & 13.35 & 0.672 & 0.502 & 15.57 & 0.601 & 0.397 \\ 
        KinD   \cite{KinD}   \scalebox{0.6}{MM'19} & S & 19.66 & 0.868 & 0.156 & 20.78 & 0.868 & 0.142 \\ 
        LLFlow     \cite{LLFlow}  \scalebox{0.6}{AAAI'22} & S &  \textcolor{blue}{ \textbf{23.68}} &  \textcolor{red}{ \textbf{0.916}} &  \textcolor{blue}{\textbf{0.125}} & 26.61 &  \textcolor{red}{ \textbf{0.924}} &  \textcolor{blue}{ \textbf{0.090}} \\ 
        LLFormer   \cite{LLFormer}   \scalebox{0.6}{AAAI'23} & S & 23.35 & 0.873 & 0.168 &  \textcolor{blue}{ \textbf{27.42}} & 0.887 & 0.144 \\ \hline
        DRBN   \cite{DRBN}  \scalebox{0.6}{CVPR'20} & SE & 18.76 & 0.880 & 0.162 & 20.14 & 0.875 & 0.149 \\ 
        LMT-GP (Ours) & SE &  \textcolor{red}{\textbf{24.12}} &  \textcolor{blue}{\textbf{0.902}} &  \textcolor{red}{\textbf{0.107}} &  \textcolor{red}{ \textbf{28.55}} &  \textcolor{blue}{ \textbf{0.921}} &  \textcolor{red}{ \textbf{0.071}} \\ \bottomrule
    \end{tabular}
\label{tab:lol_}
\end{table}
\vspace{-1.4cm}
\vspace{-0.5cm}
\noindent EnGAN, SCI, and RUAS are dark. The results of ZeroDCE++, MBLLEN, DeepUPE, and RetinexDIP suffer from varying degrees of color distortion while RetinexNet, PairLIE, KinD, GLADNet, and DRBN produce blurry images with noise. The LLFlow and LLFormer algorithms produce artifacts at the boundaries of the annotations. In contrast, LMT-GP outperforms the other methods in terms of the visual quality when compared to the GT.

\begin{figure}[H]
    \centering
    \begin{subfigure}{0.15\linewidth}
    \includegraphics[width=\linewidth]{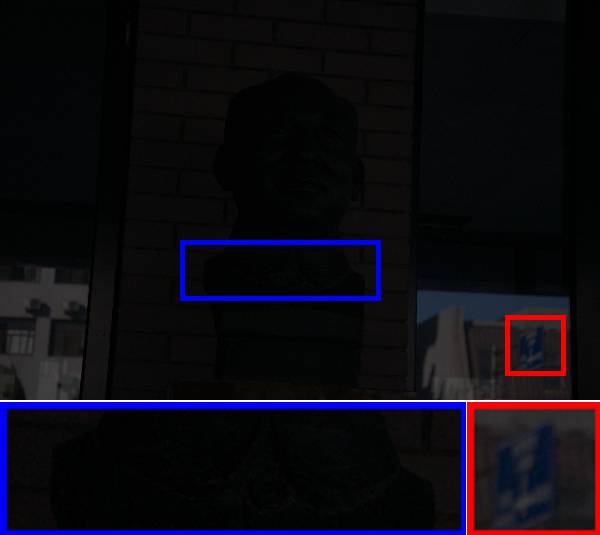}
    \subcaption*{\centering{Input}}
    \label{fig:enter-label1}
    \end{subfigure}
     \hfill
    \begin{subfigure}{0.15\linewidth}
    \includegraphics[width=\linewidth]{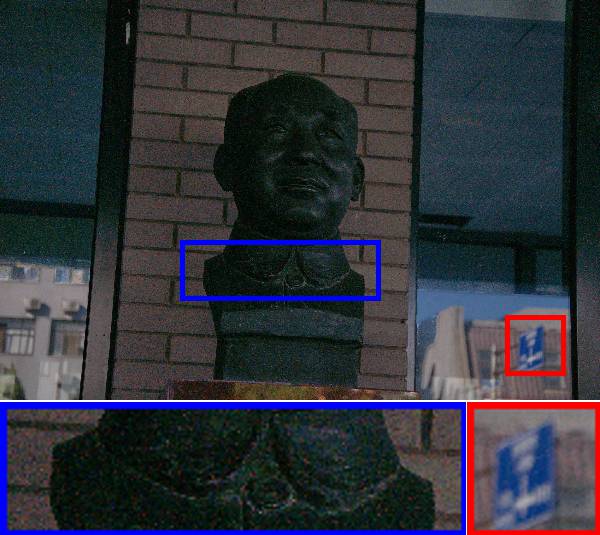}
    \subcaption*{\centering{LIME}}
    \label{fig:enter-label2}
    \end{subfigure}
     \hfill
    \begin{subfigure}{0.15\linewidth}
    \includegraphics[width=\linewidth]{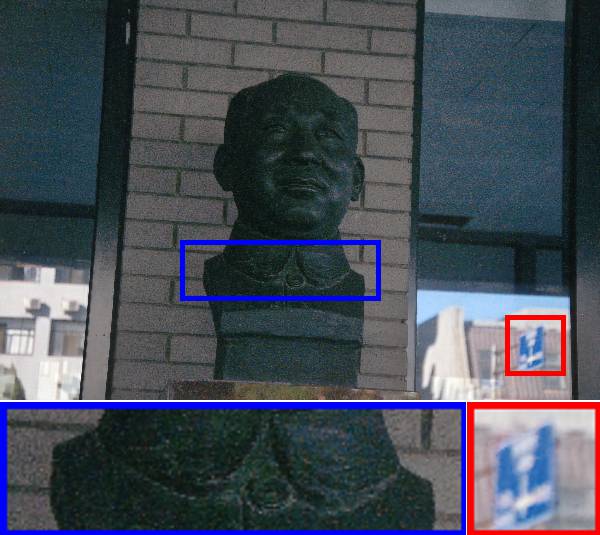}
    \subcaption*{\centering{EnGAN}}
    \label{fig:enter-label3}
    \end{subfigure}
    \hfill
\begin{subfigure}{0.15\linewidth}
    \includegraphics[width=\linewidth]{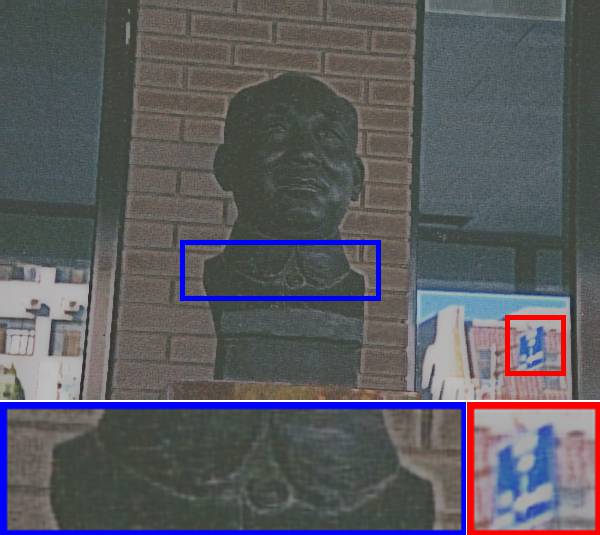}
    \subcaption*{\centering{PairLIE}}
    \label{fig:enter-label4}
    \end{subfigure}
     \hfill
    \begin{subfigure}{0.15\linewidth}
    \includegraphics[width=\linewidth]{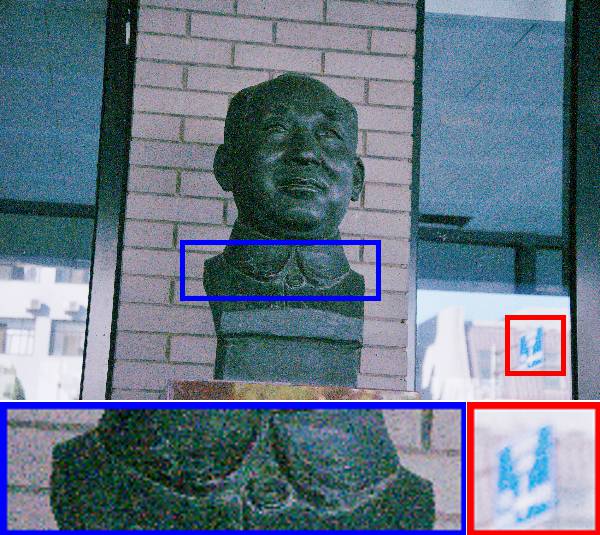}
    \subcaption*{\centering{ZeroDCE++}}
    \label{fig:enter-label6}
    \end{subfigure}
         \hfill
    \begin{subfigure}{0.15\linewidth}
    \includegraphics[width=\linewidth]{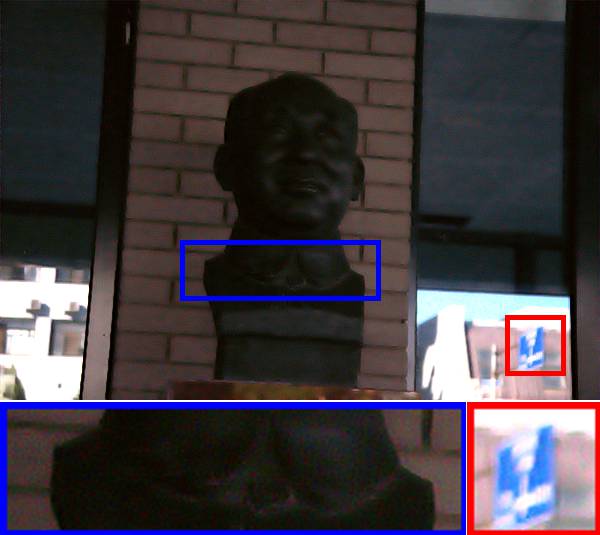}
    \subcaption*{\centering{RUAS}}
    \label{fig:enter-label7}
    \end{subfigure}
         \hfill
                 \begin{subfigure}{0.15\linewidth}
    \includegraphics[width=\linewidth]{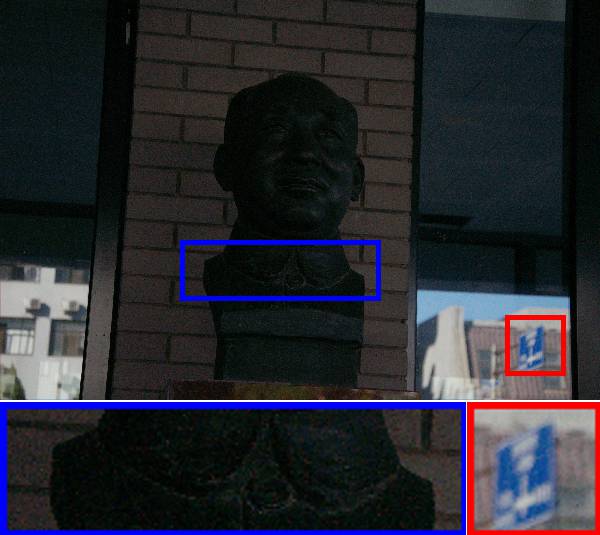}
    \subcaption*{\centering{SCI}}
    \label{fig:enter-label4}
    \end{subfigure}
     \hfill
    \begin{subfigure}{0.15\linewidth}
    \includegraphics[width=\linewidth]{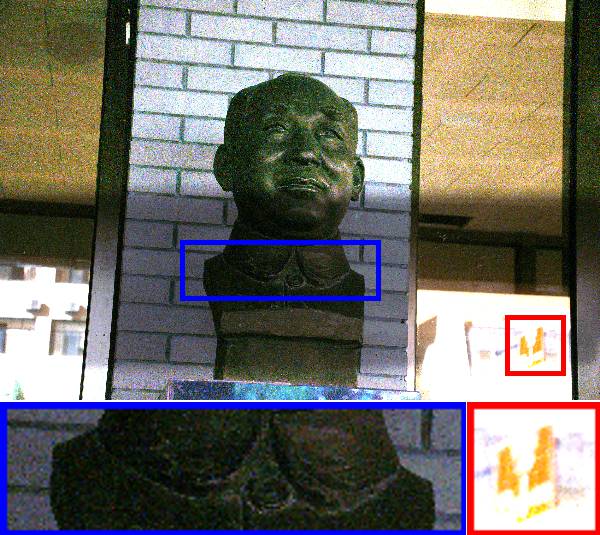}
    \subcaption*{\centering{RetinexDIP}}
    \label{fig:enter-label8}
    \end{subfigure}
         \hfill
    \begin{subfigure}{0.15\linewidth}
    \includegraphics[width=\linewidth]{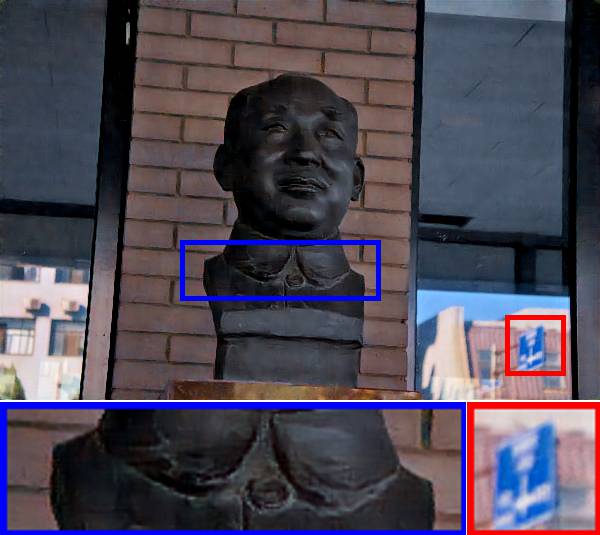}
    \subcaption*{\centering{MBLLEN}}
    \label{fig:enter-label9}
    \end{subfigure}
         \hfill
    \begin{subfigure}{0.15\linewidth}
    \includegraphics[width=\linewidth]{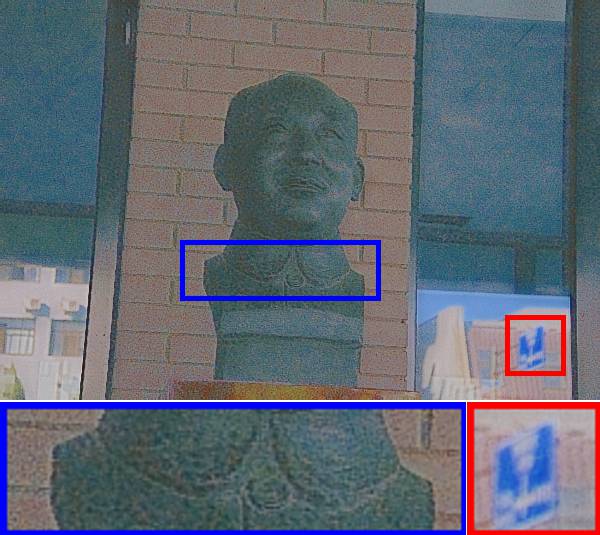}
    \subcaption*{\centering{RetinexNet}}
    \label{fig:enter-label10}
    \end{subfigure}
         \hfill
    \begin{subfigure}{0.15\linewidth}
    \includegraphics[width=\linewidth]{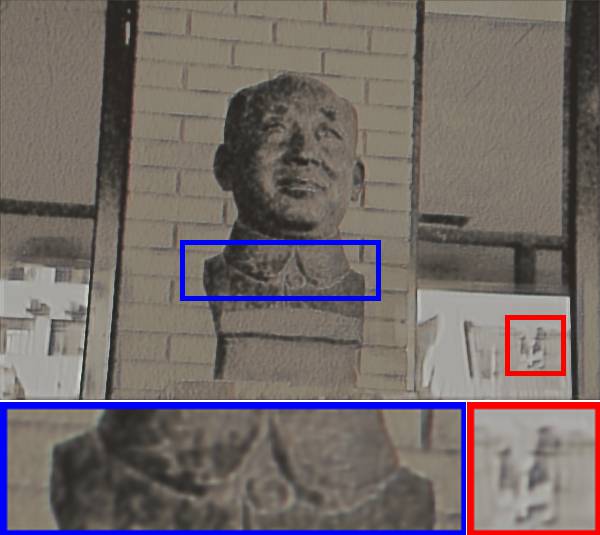}
    \subcaption*{\centering{GLADNet}}
    \label{fig:enter-label11}
    \end{subfigure}
         \hfill
    \begin{subfigure}{0.15\linewidth}
    \includegraphics[width=\linewidth]{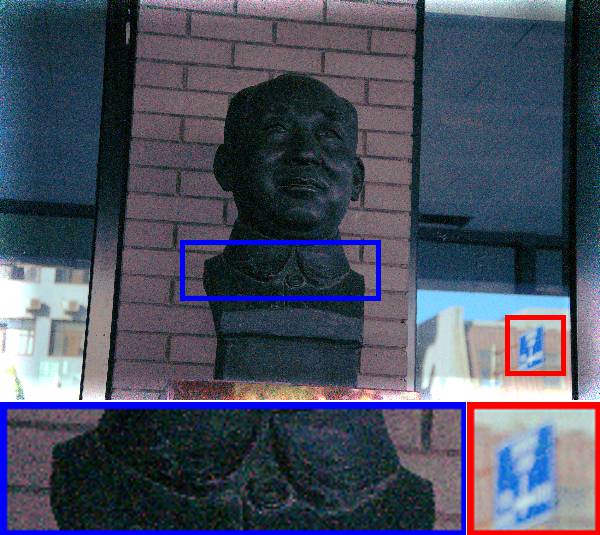}
    \subcaption*{\centering{DeepUPE}}
    \label{fig:enter-label12}
    \end{subfigure}
         \hfill
    \begin{subfigure}{0.15\linewidth}
    \includegraphics[width=\linewidth]{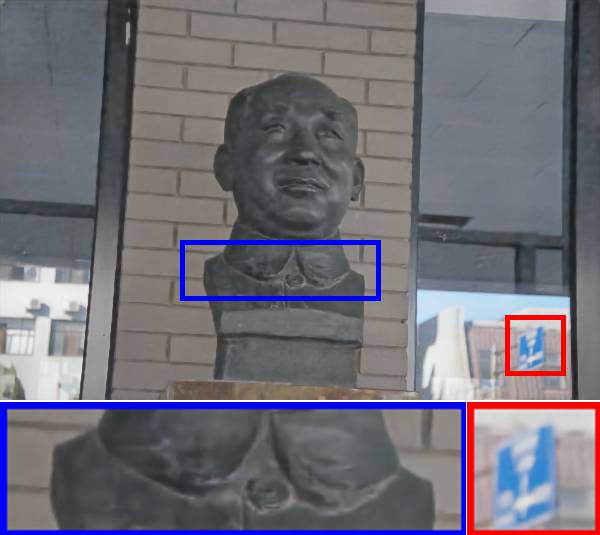}
    \subcaption*{\centering{KinD}}
    \label{fig:enter-label13}
    \end{subfigure}
         \hfill
    \begin{subfigure}{0.15\linewidth}
    \includegraphics[width=\linewidth]{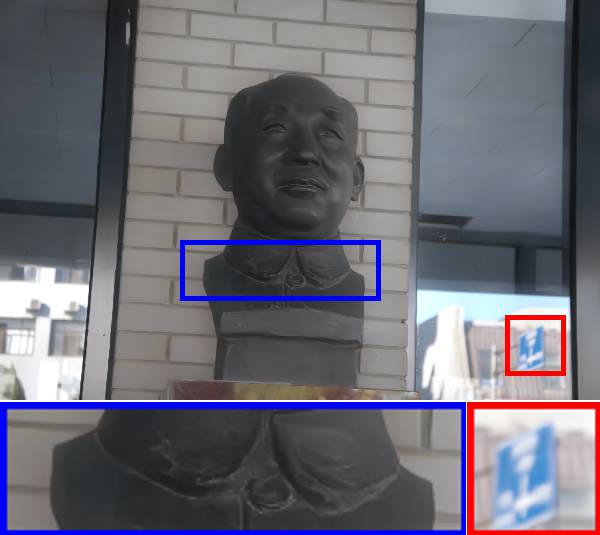}
    \subcaption*{\centering{LLFlow}}
    \label{fig:enter-label14}
    \end{subfigure}
         \hfill
    \begin{subfigure}{0.15\linewidth}
    \includegraphics[width=\linewidth]{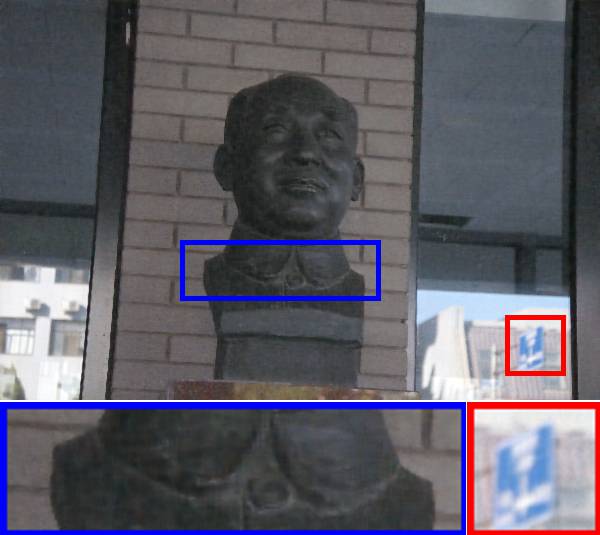}
    \subcaption*{\centering{LLFormer}}
    \label{fig:enter-label15}
    \end{subfigure}
         \hfill
    \begin{subfigure}{0.15\linewidth}
    \includegraphics[width=\linewidth]{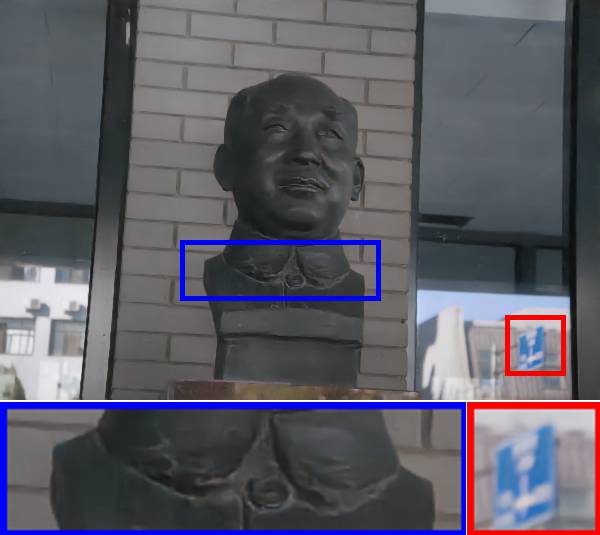}
    \subcaption*{\centering{DRBN}}
    \label{fig:enter-label16}
    \end{subfigure}
         \hfill
    \begin{subfigure}{0.15\linewidth}
    \includegraphics[width=\linewidth]{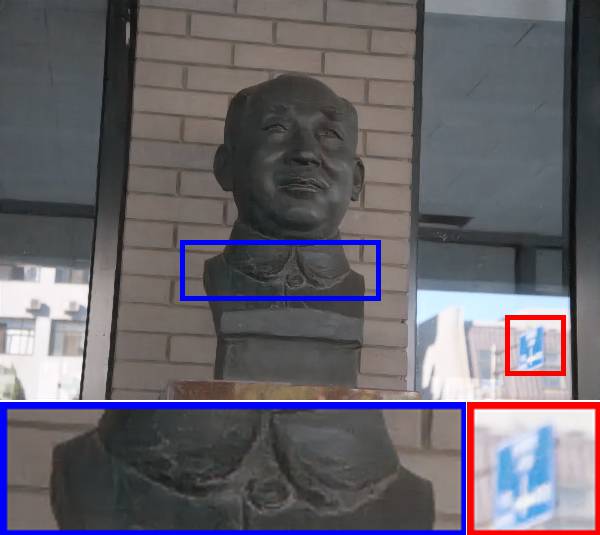}
    \subcaption*{\centering{Ours}}
    \label{fig:enter-label17}
    \end{subfigure}
         \hfill
    \begin{subfigure}{0.15\linewidth}
    \includegraphics[width=\linewidth]{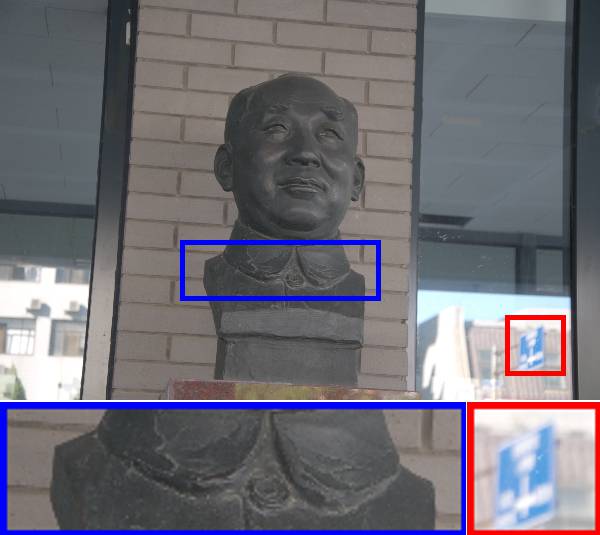}
    \subcaption*{\centering{GT}}
    \label{fig:enter-label18}
    \end{subfigure}
\caption{Visual comparison of state-of-the-art LLIE methods on the VE-LOL dataset. More results can be found in the Supplementary Materials.}
    \label{fig:velol}
\end{figure}

\noindent\textbf{Evaluation on VE-LOL. }LMT-GP outperforms the second-best method on the VE-LOL dataset in terms of PSNR and LPIPS scores by 1.13 dB and 0.019, respectively, as shown in \Cref{tab:lol_}. However, its SSIM score is slightly lower than that of the LLFlow. Meanwhile, as shown in \cref{fig:velol}, the previous methods can not adequately enhance lighting or maintain details in the annotated regions. 
\vspace{-1.2cm}

\begin{table}[H]\scriptsize
\caption{Quantitative results of face detection on the Darkface dataset. ``Original” means no pre-processing measures. The best result is in \textcolor{red}{\textbf{red}} whereas the second-best is in \textcolor{blue}{\textbf{blue}}.}
\renewcommand{\arraystretch}{1.2}
\centering
\begin{tabular}{c|ccccc} 
\toprule
\makebox[0.15\textwidth][c]{Method}                                                        & \makebox[0.14\textwidth][c]{Original} & \makebox[0.14\textwidth][c]{EnGAN}  & \makebox[0.15\textwidth][c]{PairLIE}  & \makebox[0.15\textwidth][c]{MBLLEN} & \makebox[0.15\textwidth][c]{RetinexNet}        
\\
\hline
mAP (\%)   
& 40.47    & 37.34  & 40.97    & 42.98  & 41.38                             \\ 
\hline\hline
\multicolumn{1}{c}{GLADNet}                                   & KinD     & LLFlow & LLFormer & DRBN   & Ours                              \\
\hline
\multicolumn{1}{c}{\textbf{\textbf{\textcolor{blue}{43.73}}}} & 42.36    & 41.26  & 42.81    & 42.70  & \textcolor{red}{\textbf{43.92 }}  \\
\bottomrule
\end{tabular}
\label{tab:darkface}
\end{table}
\vspace{-1cm}

\begin{figure}[H]
     \centering
    \begin{subfigure}{0.15\linewidth}
    \includegraphics[width=\linewidth]{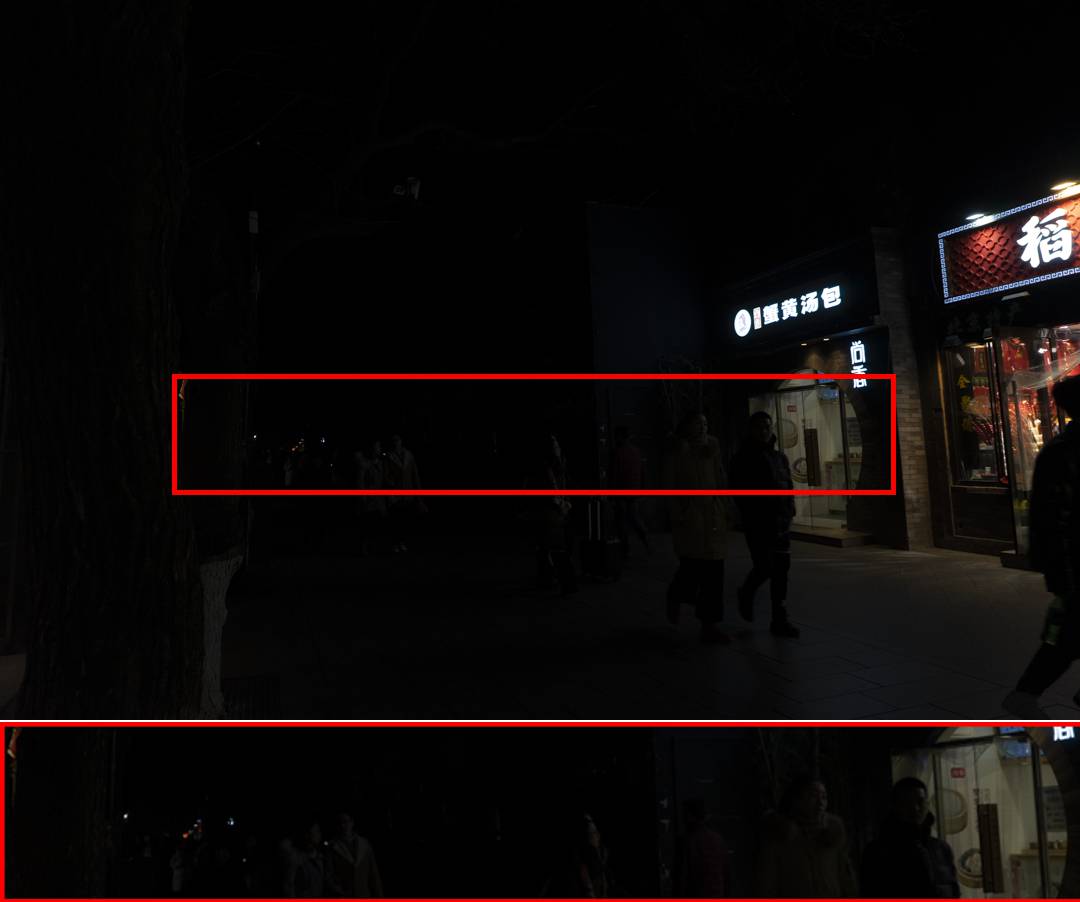}
    \subcaption*{\centering{Input}}
    \label{fig:enter-label1}
    \end{subfigure}
     \hfill
         \begin{subfigure}{0.15\linewidth}
    \includegraphics[width=\linewidth]{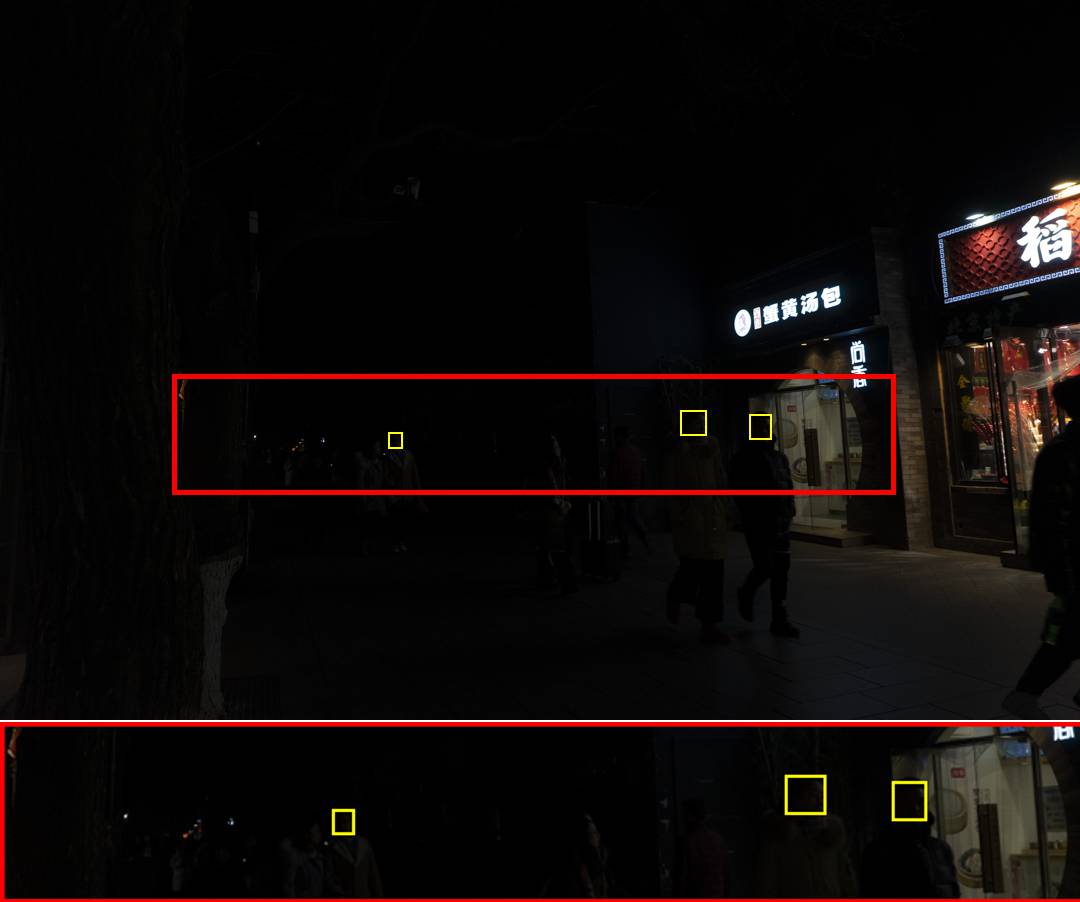}
    \subcaption*{\centering{None}}
    \label{fig:enter-label1}
    \end{subfigure}
     \hfill
    \begin{subfigure}{0.15\linewidth}
    \includegraphics[width=\linewidth]{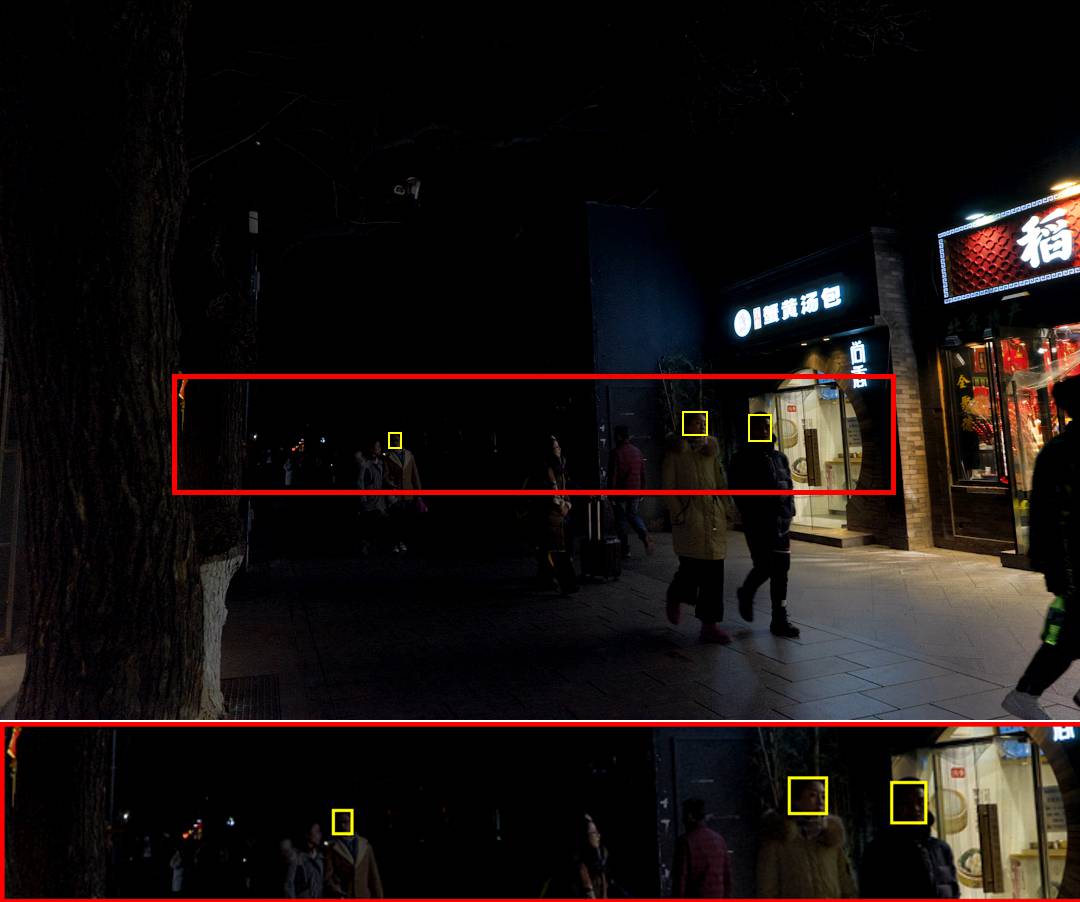}
    \subcaption*{\centering{EnGAN}}
    \label{fig:enter-label2}
    \end{subfigure}
     \hfill
    \begin{subfigure}{0.15\linewidth}
    \includegraphics[width=\linewidth]{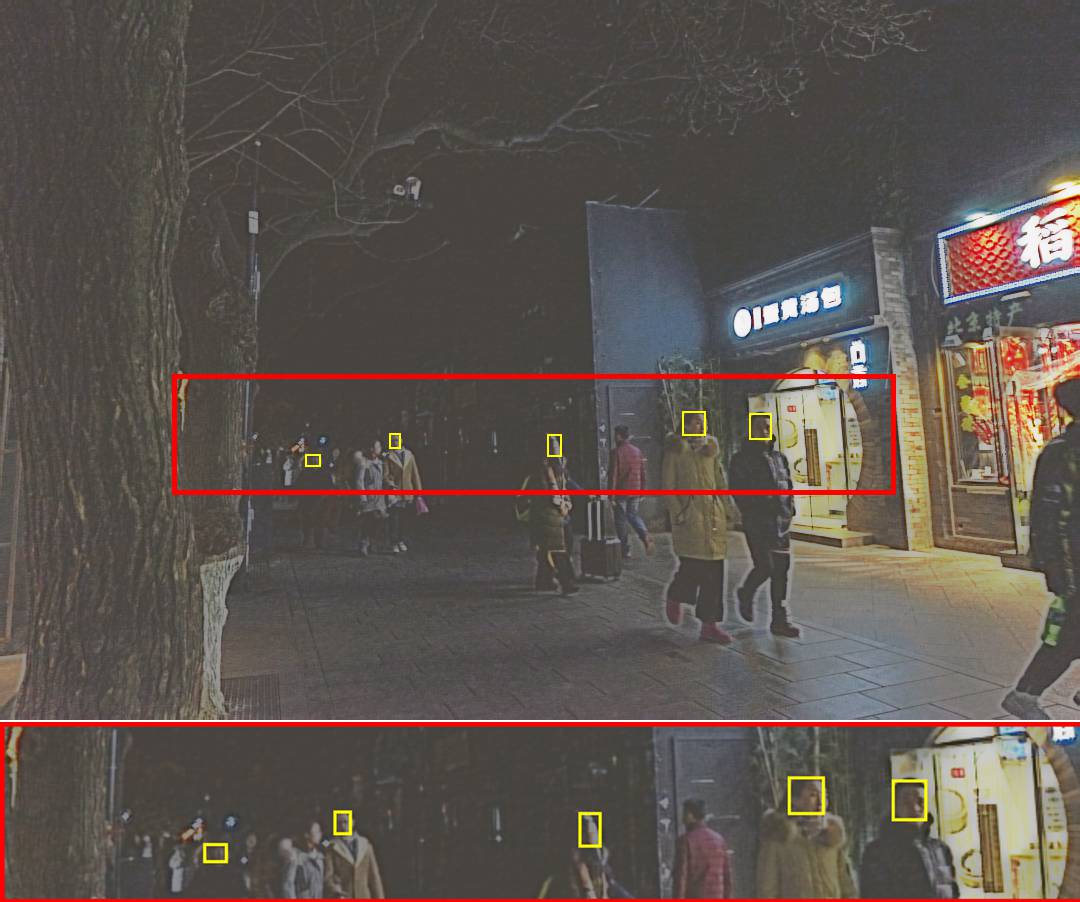}
    \subcaption*{\centering{PairLIE}}
    \label{fig:enter-label3}
    \end{subfigure}
    \hfill
        \begin{subfigure}{0.15\linewidth}
    \includegraphics[width=\linewidth]{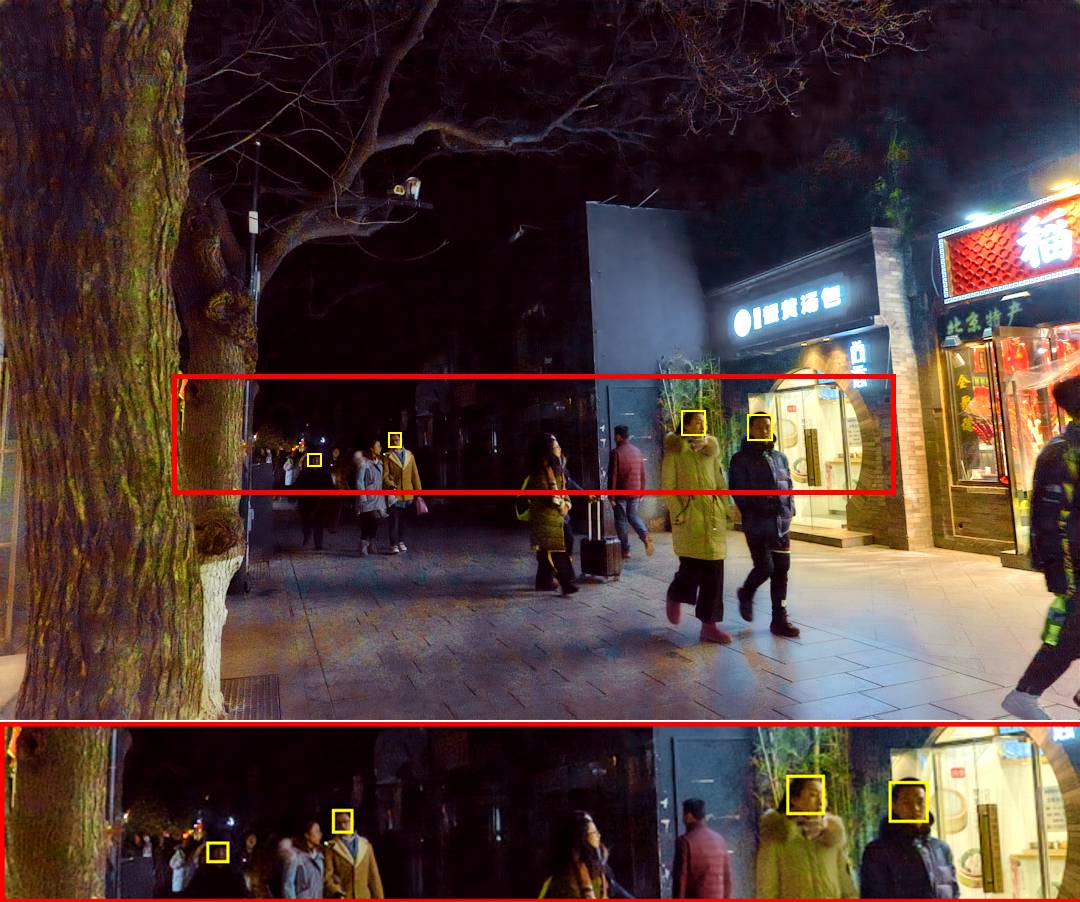}
    \subcaption*{\centering{MBLLEN}}
    \label{fig:enter-label4}
    \end{subfigure}
     \hfill
              \begin{subfigure}{0.15\linewidth}
    \includegraphics[width=\linewidth]{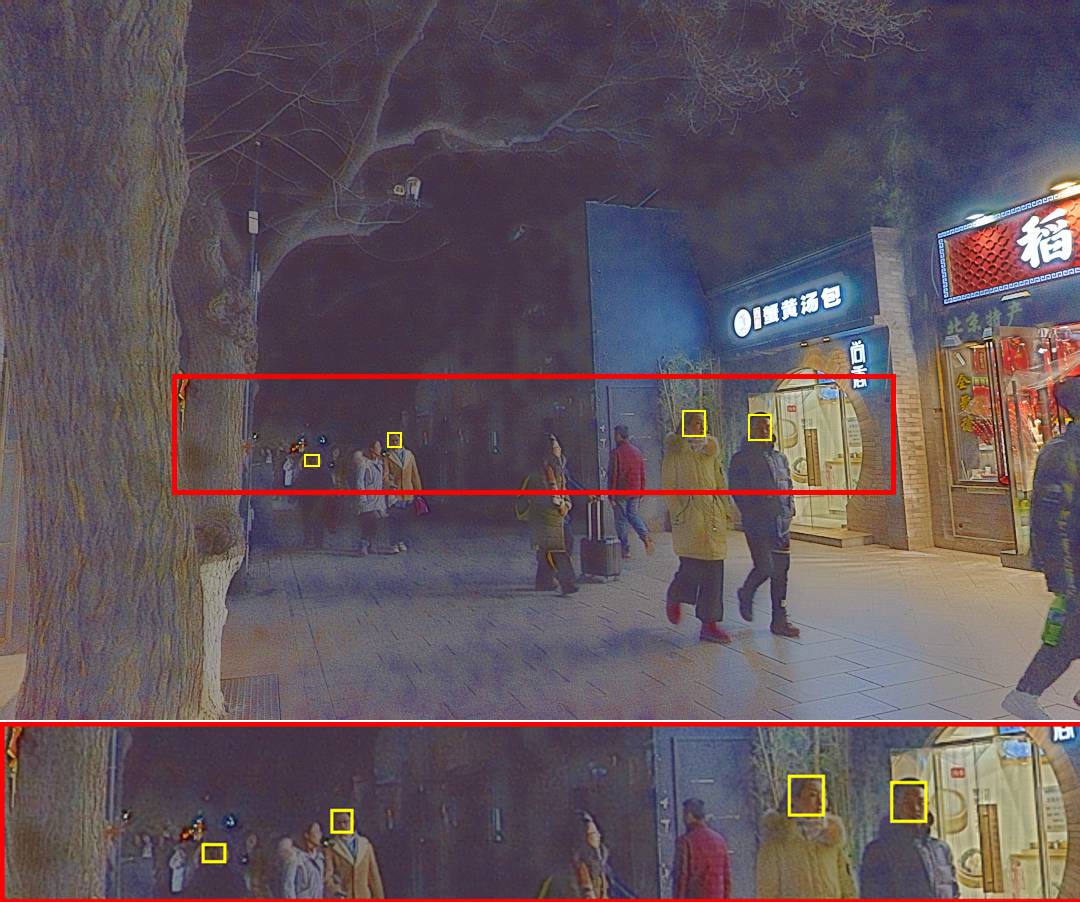}
    \subcaption*{\centering{RetinexNet}}
    \label{fig:enter-label5}
    \end{subfigure}
     \hfill
    \begin{subfigure}{0.15\linewidth}
    \includegraphics[width=\linewidth]{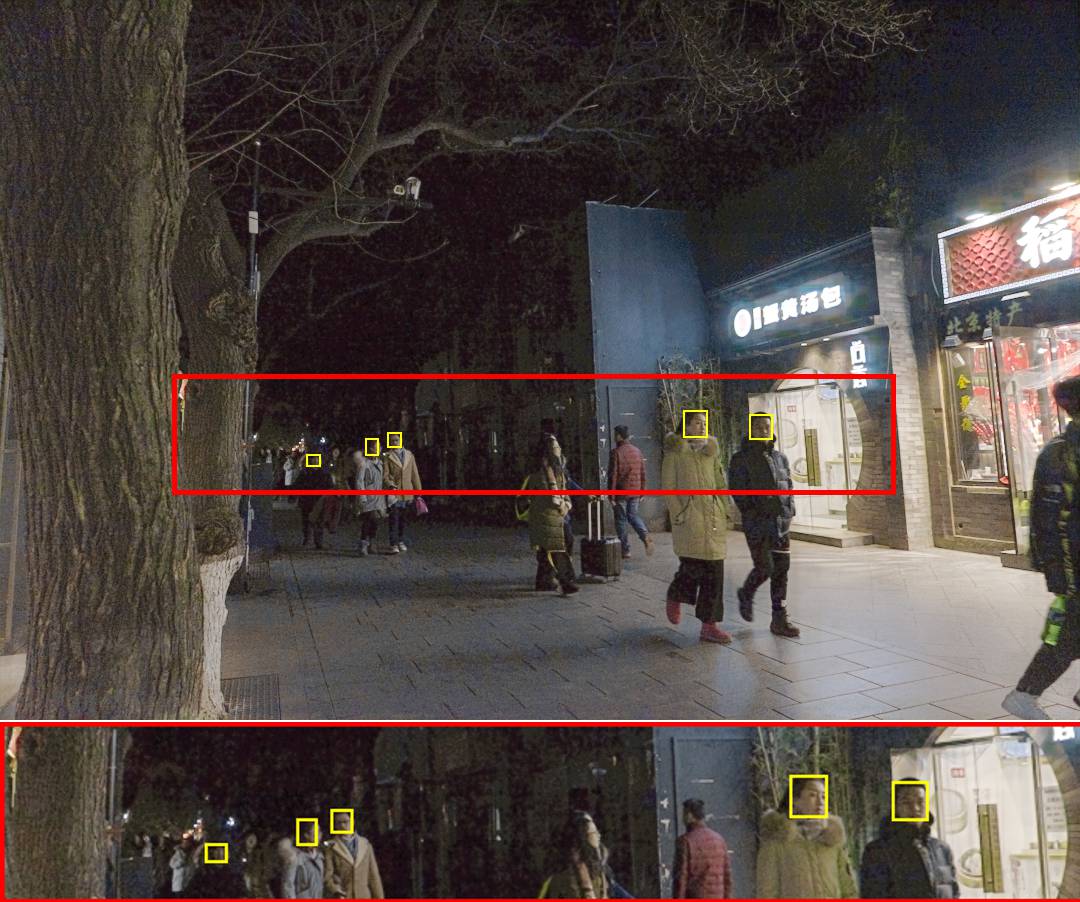}
    \subcaption*{\centering{GLADNet}}
    \label{fig:enter-label6}
    \end{subfigure}
         \hfill
    \begin{subfigure}{0.15\linewidth}
    \includegraphics[width=\linewidth]{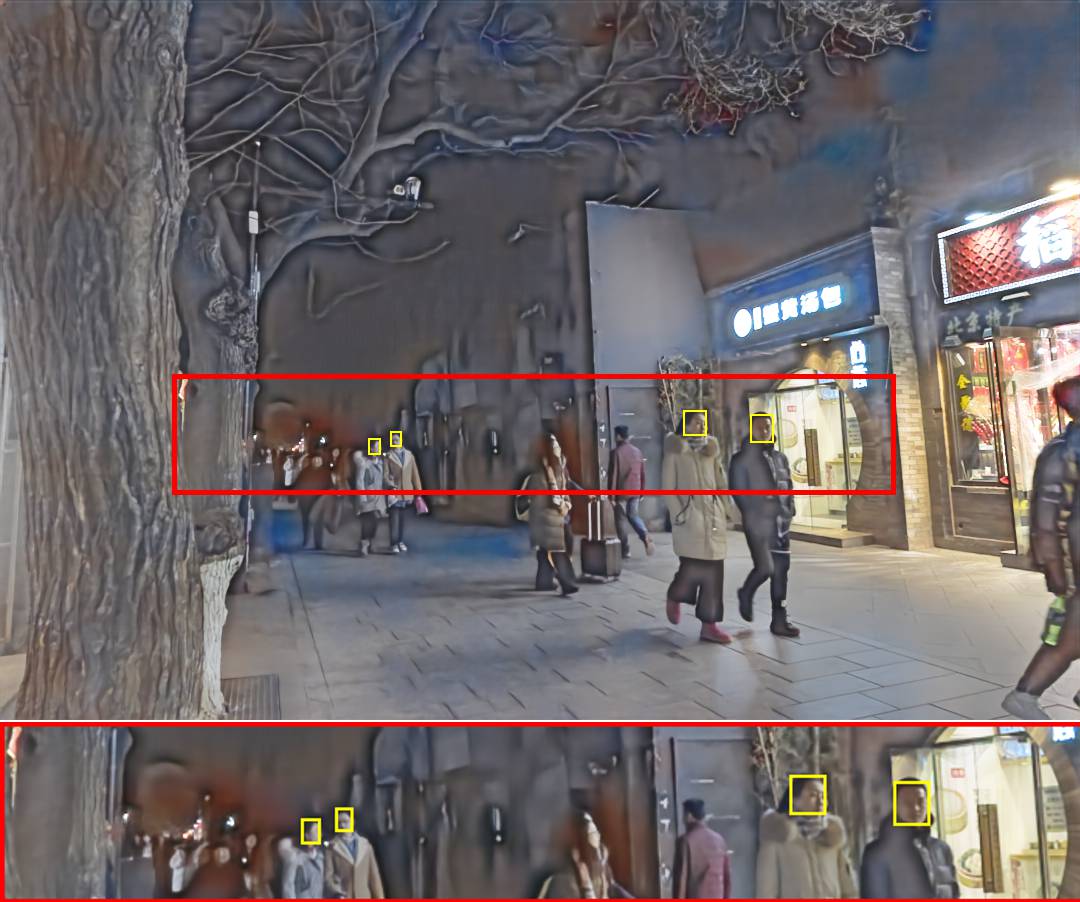}
    \subcaption*{\centering{KinD}}
    \label{fig:enter-label7}
    \end{subfigure}
         \hfill
    \begin{subfigure}{0.15\linewidth}
    \includegraphics[width=\linewidth]{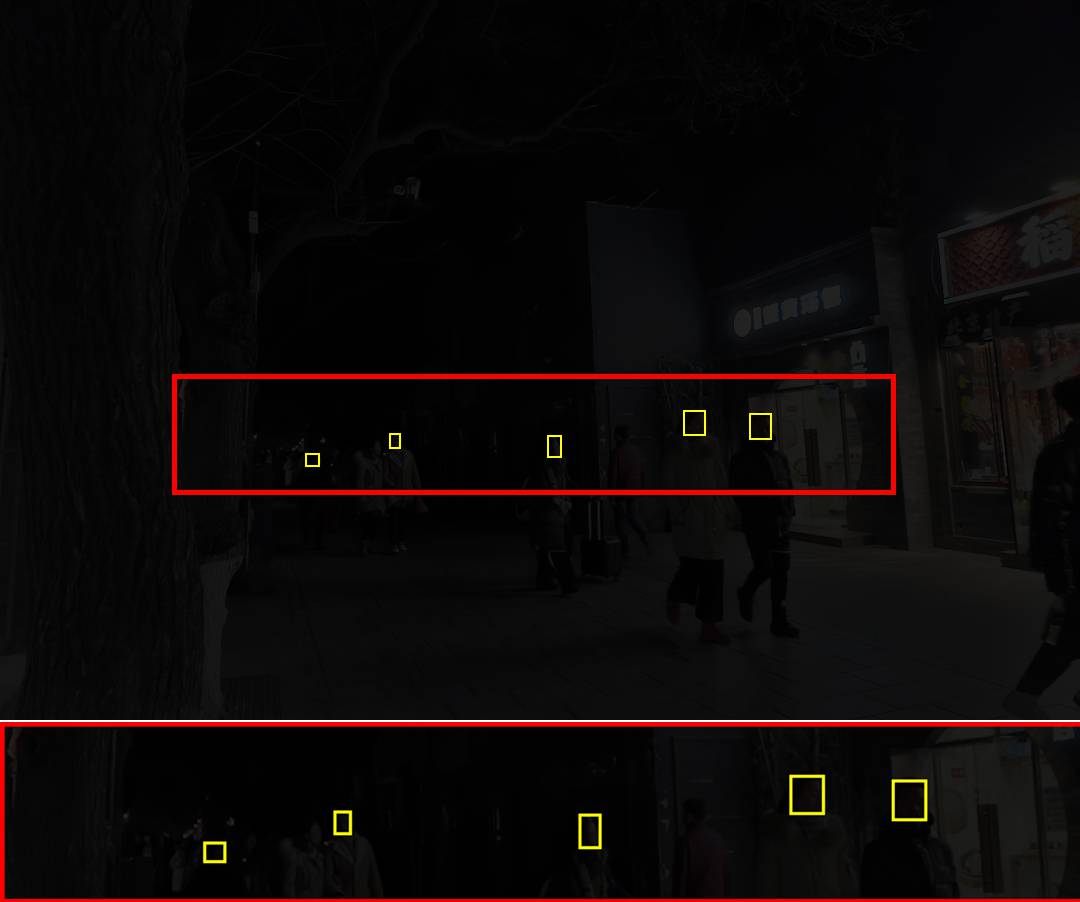}
    \subcaption*{\centering{LLFlow}}
    \label{fig:enter-label8}
    \end{subfigure}
         \hfill
    \begin{subfigure}{0.15\linewidth}
    \includegraphics[width=\linewidth]{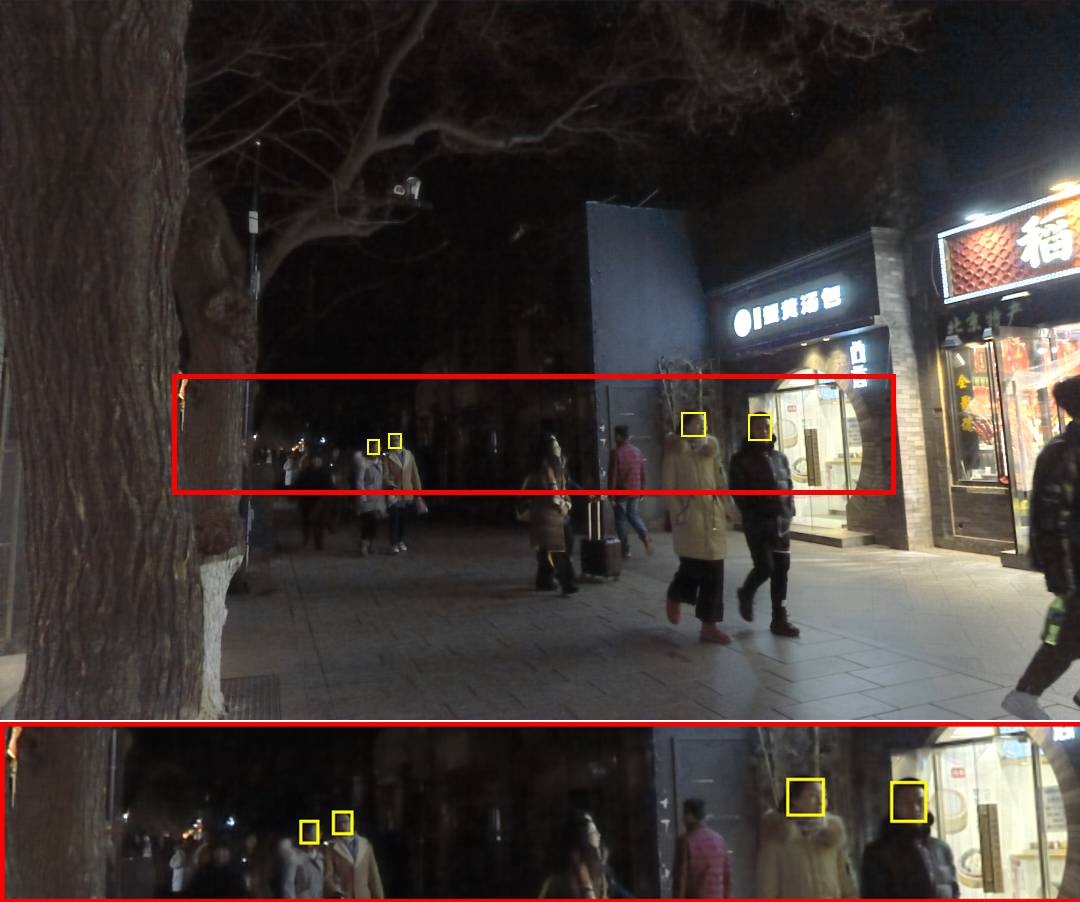}
    \subcaption*{\centering{LLFormer}}
    \label{fig:enter-label9}
    \end{subfigure}
         \hfill
    \begin{subfigure}{0.15\linewidth}
    \includegraphics[width=\linewidth]{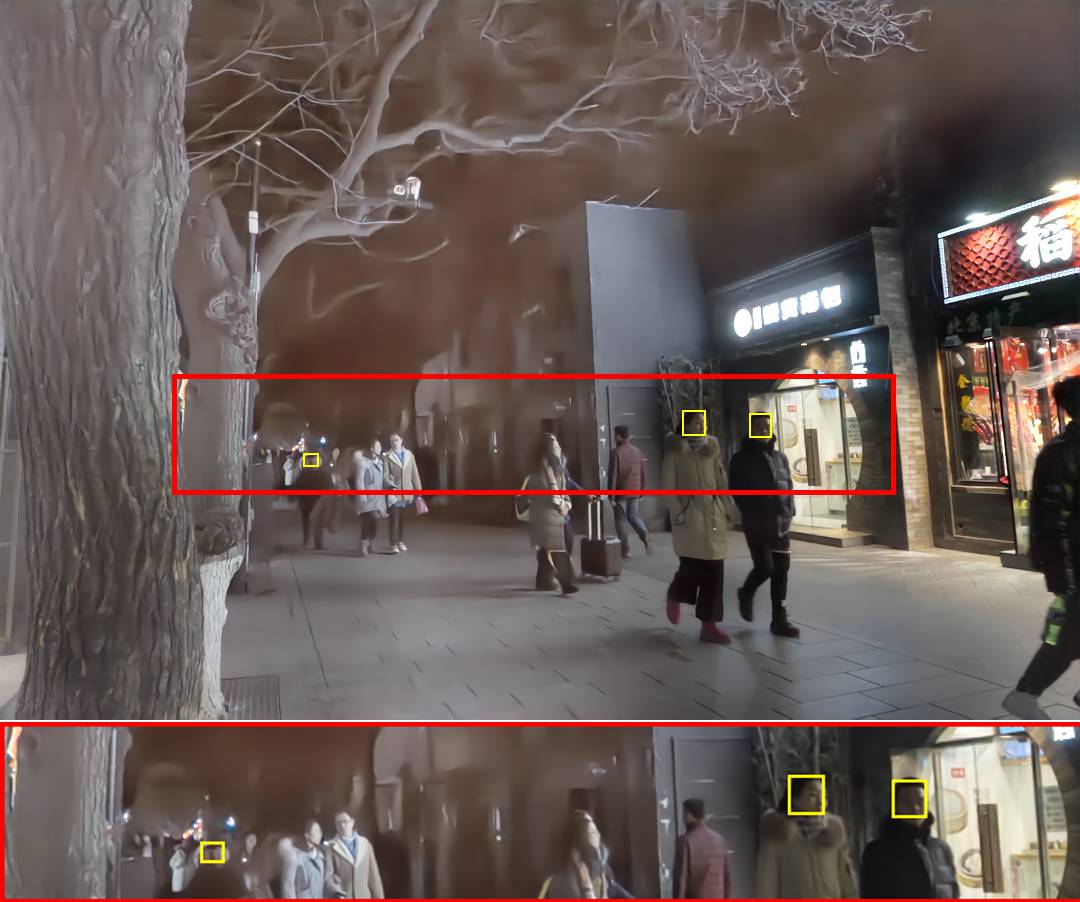}
    \subcaption*{\centering{DRBN}}
    \label{fig:enter-label10}
    \end{subfigure}
         \hfill
    \begin{subfigure}{0.15\linewidth}
    \includegraphics[width=\linewidth]{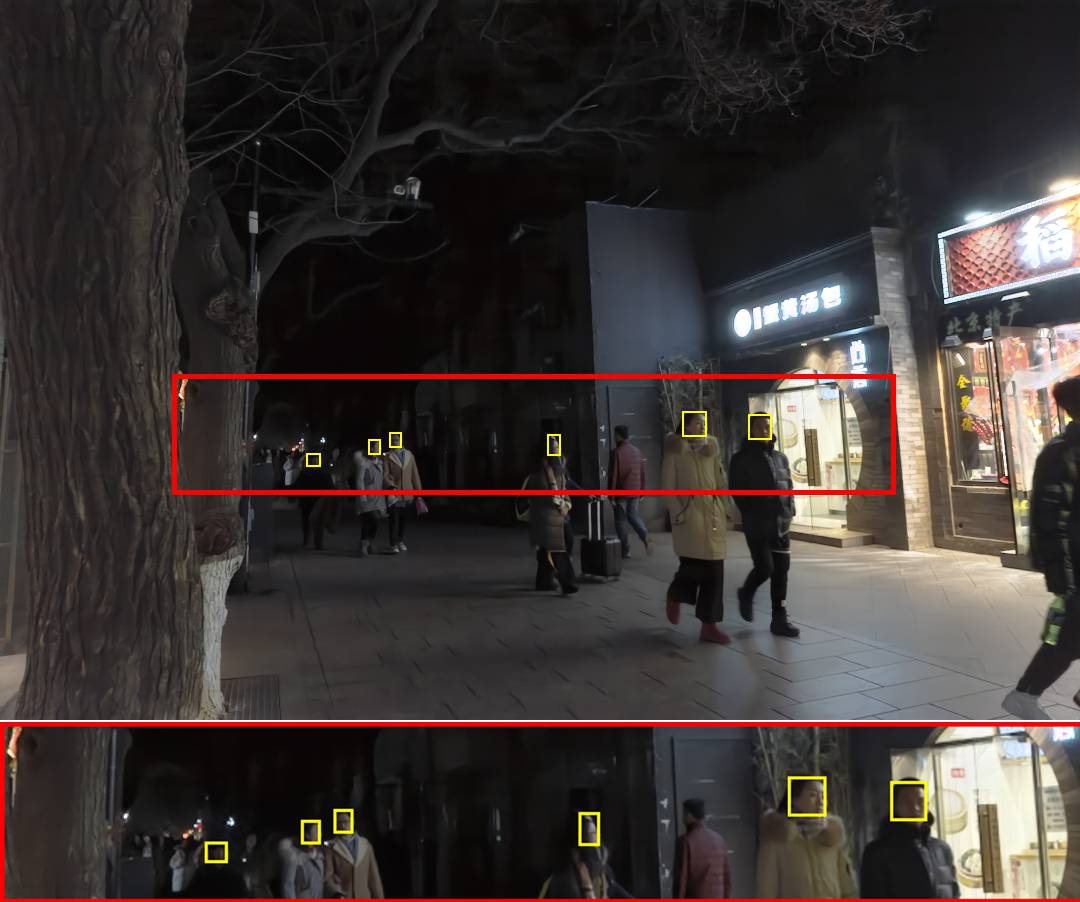}
    \subcaption*{\centering{Ours}}
    \label{fig:enter-label11}
    \end{subfigure}
         \hfill
\caption{Visual comparison of face detection on the Darkface dataset. More results can be found in the Supplementary Materials.}
\label{fig:darkface}
\end{figure}
\vspace{-0.5cm}

\noindent However, our method successfully reduces noise, brighten images, and preserves fine texture details due to the use of latent vectors to learn low-level features.

\noindent\textbf{Evaluation on Darkface. }As shown in \Cref{tab:darkface}, LMT-GP yields the highest mAP score. It outerforms the second-best method by $0.19\%$ and surpasses the Original by $3.45\%$. The qualitative results are shown in \cref{fig:darkface}. LMT-GP method
\begin{table}[H]\scriptsize
\caption{Quantitative results of nighttime segmentation performance on the ACDC dataset. ``Original" means no pre-processing measures. The symbol set $\{$RO, SI, BU, WA, FE, PO, LI, SI, VE, TE, SK, PE, CA, TR, BI$\}$ represents $\{$road, sidewalk, building, wall, fence, pole, light, sign, vegetation, terrain, sky, person, car, train, bicycle$\}$. The values for the categories ``rider”, ``truck”, ``bus”, and ``motorcycle” are all zero, therefore they are not displayed in the table. The best results are in \textcolor{red}{\textbf{red}} whereas the second-best are in \textcolor{blue}{\textbf{blue}}.}
\renewcommand{\arraystretch}{1.1}
\centering
\begin{tabular}{l|cccccccc} 
\toprule
\multicolumn{1}{c|}{\makebox[0.1\textwidth][c]{Method}} & \makebox[0.1\textwidth][c]{RO}  & \makebox[0.1\textwidth][c]{SI}  & \makebox[0.1\textwidth][c]{BU}  & \makebox[0.1\textwidth][c]{WA}  & \makebox[0.1\textwidth][c]{FE}  & \makebox[0.1\textwidth][c]{PO}  & \makebox[0.1\textwidth][c]{LI}  & \makebox[0.1\textwidth][c]{SI}  \\ 
\hline
Original  & 89.77  & 60.42  & 69.34  & 33.05  & 29.95  & 35.29  & 12.00   & 17.57  \\
RetinexNet  & 90.62  & 61.82  & 67.58  & 37.15  & 28.51  & 35.34  & 5.45    & 9.69   \\
KinD  & 90.85  & 63.62  & 68.89  & 34.49  & 31.38  & 35.33  & 23.26   & 15.19  \\
EnGAN    & 89.12  & 59.33  & 70.14  & 35.98  & 31.78  & 37.85  & 16.51   & 13.95  \\
GLADNet  & 88.12  & 57.47  & 64.78  & 25.72  & 26.63  & 32.16  & 3.78    & 10.83  \\
MBLLEN   & 90.18  & 60.92  & 68.53  & 32.39  & 31.28  & 34.95  & 2.17    & 11.99  \\
LLFlow   & 90.67  & 63.61  & 68.82  & 33.56  & 31.02  & 35.49  & 19.86   & 10.99  \\
LLFormer  & 90.25  & 61.32  & 69.55  & \textcolor{blue}{\textbf{37.54}}   & 30.42  & 33.03  & 8.62    & 11.96  \\
PairLIE  & 89.99  & 62.15  & \textbf{\textcolor{blue}{70.63}}   & \textcolor{red}{\textbf{39.29}}    & \textbf{\textcolor{red}{35.91}}    & \textbf{\textcolor{blue}{39.19}}   & \textbf{\textcolor{blue}{39.39}} & \textbf{\textcolor{blue}{19.35}}  \\
DRBN  & \textbf{\textcolor{blue}{91.08}}   & \textbf{\textcolor{blue}{64.23}}  & 70.54  & 37.01  & \textcolor{blue}{\textbf{32.15}}   & 36.81  & 31.87   & 16.55  \\
Ours  & \textbf{\textcolor{red}{92.05}}    & \textbf{\textcolor{red}{67.17}}   & \textbf{\textcolor{red}{70.79}}    & 37.32  & 30.32  & \textcolor{red}{\textbf{41.22}}    & \textbf{\textcolor{red}{44.85}}  & \textcolor{red}{\textbf{22.04}}   \\ 
\hline\hline
Method   & VE  & TE  & SK  & PE  & CA  & TR  & BI  & mIOU\\
\hline
Original  & 52.15  & 6.38  & 71.03  & 1.15 & 46.78  & 76.70  & 0.0  & 31.66 \\
RetinexNet  & 56.85  & 5.25  & 70.74  & 1.00 & 47.09  & 84.23  & 0.0  & 31.65 \\
KinD  & 58.32  & 9.29  & 72.11  & 12.58  & 49.38  & 82.73  & 0.0  & 34.07 \\
EnGAN    & \textbf{\textbf{\textcolor{red}{59.30}}}  & 5.76  & 72.12  & 10.60  & 45.84  & 81.25  & 0.0  & 33.13 \\
GLADNet  & 52.11  & 1.53  & 68.17  & 0.20 & 38.50  & 75.08  & 0.0  & 28.69 \\
MBLLEN   & 57.2 & 6.97  & 71.83  & 0.60 & 41.44  & 79.80  & 0.0  & 14.66 \\
LLFlow   & \textbf{\textbf{\textcolor{blue}{59.27}}} & 8.73  & \textbf{\textbf{\textcolor{red}{73.90}}}  & 7.81 & 44.91  & \textbf{\textbf{\textcolor{red}{85.98}}}  & 0.0  & 33.40 \\
LLFormer  & 58.77  & 6.46  & 71.44  & 2.53 & 44.76  & 79.63  & 0.0  & 31.91 \\
PairLIE  & 57.79  & 4.62  & 71.81  & \textbf{\textbf{\textcolor{blue}{18.83}}} & \textbf{\textbf{\textcolor{blue}{50.55}}} & \textbf{\textbf{\textcolor{blue}{85.30}}} & \textbf{\textcolor{blue}{1.29}}  & \textbf{\textbf{\textcolor{blue}{36.11}}}  \\
DRBN  & 59.24  & \textbf{\textbf{\textcolor{blue}{9.49}}} & \textbf{\textbf{\textcolor{blue}{73.28}}} & 13.45  & 49.20  & 80.54  & 0.0  & 35.02 \\
Ours  & 58.03  & \textbf{\textbf{\textcolor{red}{13.42}}} & 72.67  & \textbf{\textbf{\textcolor{red}{23.71}}}  & \textbf{\textbf{\textcolor{red}{55.94}}} & 84.69  & \textbf{\textbf{\textcolor{red}{33.09}}}   & \textbf{\textbf{\textcolor{red}{39.33}}}  \\
\bottomrule
\end{tabular}
\label{tab:acdc}
\vspace{-0.4cm}
\end{table}
\noindent leads to more satisfactory visualization results, particularly in terms of illumination adjustment levels. We observe that more objects are accurately detected when using LMT-GP.

\begin{figure}[H]
    \centering
    \begin{subfigure}{0.155\linewidth}
    \includegraphics[width=\linewidth]{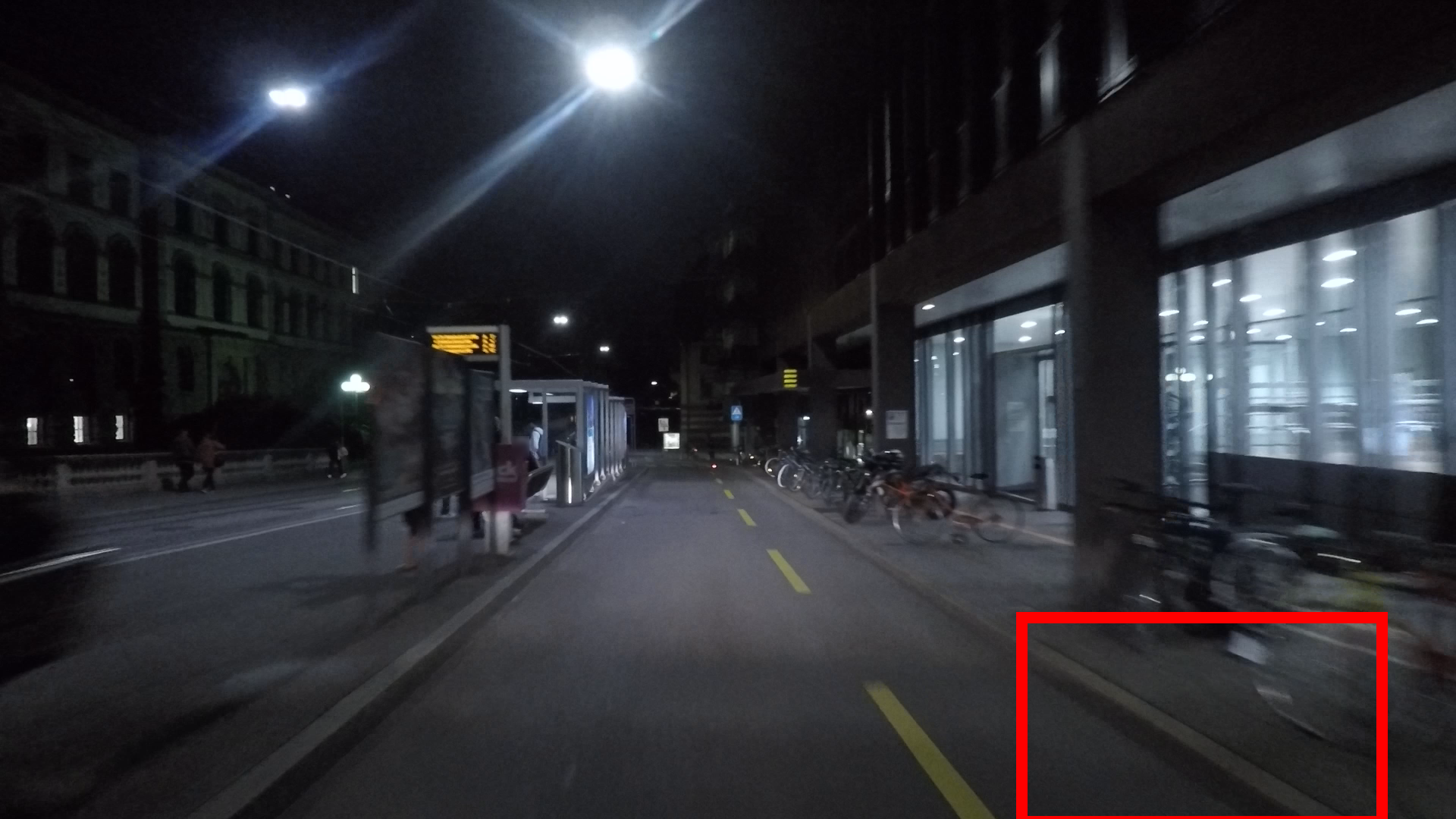}
    \subcaption*{\centering{Input}}
    \label{fig:enter-label1}
    \end{subfigure}
     \hfill
    \begin{subfigure}{0.155\linewidth}
    \includegraphics[width=\linewidth]{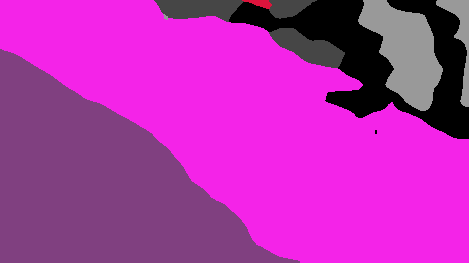}
    \subcaption*{\centering{EnGAN}}
    \label{fig:enter-label2}
    \end{subfigure}
     \hfill
    \begin{subfigure}{0.155\linewidth}
    \includegraphics[width=\linewidth]{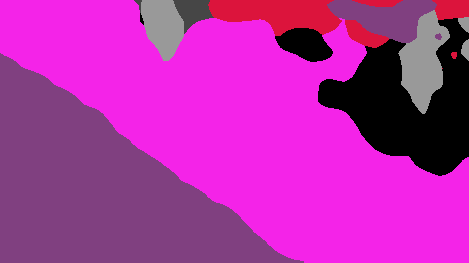}
    \subcaption*{\centering{PairLIE}}
    \label{fig:enter-label3}
    \end{subfigure}
    \hfill
        \begin{subfigure}{0.155\linewidth}
    \includegraphics[width=\linewidth]{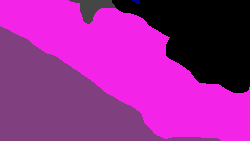}
    \subcaption*{\centering{MBLLEN}}
    \label{fig:enter-label4}
    \end{subfigure}
     \hfill
              \begin{subfigure}{0.155\linewidth}
    \includegraphics[width=\linewidth]{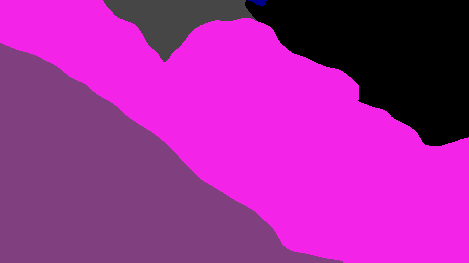}
    \subcaption*{\centering{RetinexNet}}
    \label{fig:enter-label5}
    \end{subfigure}
     \hfill
    \begin{subfigure}{0.155\linewidth}
    \includegraphics[width=\linewidth]{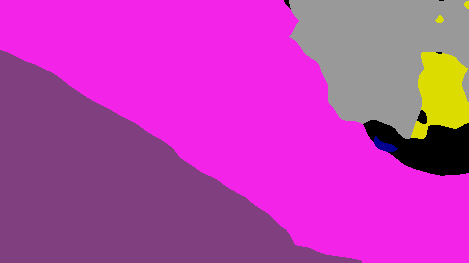}
    \subcaption*{\centering{GLADNet}}
    \label{fig:enter-label6}
    \end{subfigure}
         \hfill
    \begin{subfigure}{0.155\linewidth}
    \includegraphics[width=\linewidth]{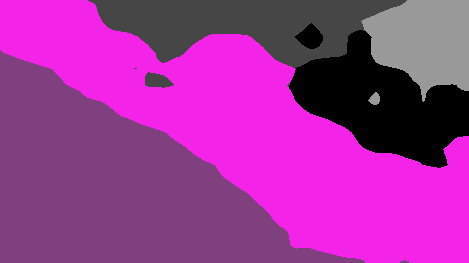}
    \subcaption*{\centering{KinD}}
    \label{fig:enter-label7}
    \end{subfigure}
         \hfill
    \begin{subfigure}{0.155\linewidth}
    \includegraphics[width=\linewidth]{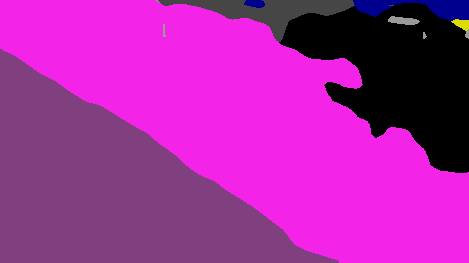}
    \subcaption*{\centering{LLFlow}}
    \label{fig:enter-label8}
    \end{subfigure}
         \hfill
    \begin{subfigure}{0.155\linewidth}
    \includegraphics[width=\linewidth]{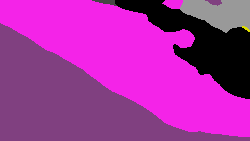}
    \subcaption*{\centering{LLFormer}}
    \label{fig:enter-label9}
    \end{subfigure}
         \hfill
    \begin{subfigure}{0.155\linewidth}
    \includegraphics[width=\linewidth]{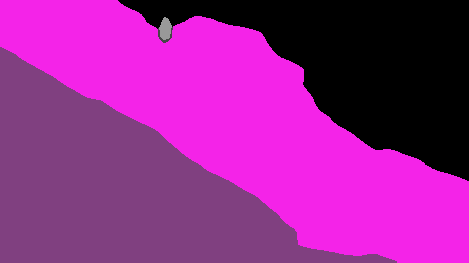}
    \subcaption*{\centering{DRBN}}
    \label{fig:enter-label10}
    \end{subfigure}
         \hfill
    \begin{subfigure}{0.155\linewidth}
    \includegraphics[width=\linewidth]{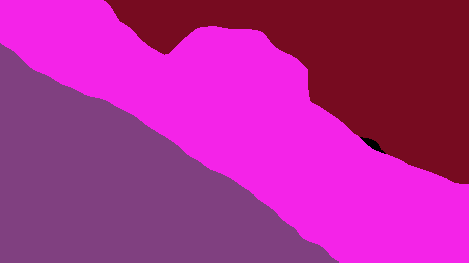}
    \subcaption*{\centering{Ours}}
    \label{fig:enter-label11}
    \end{subfigure}
         \hfill
    \begin{subfigure}{0.155\linewidth}
    \includegraphics[width=\linewidth]{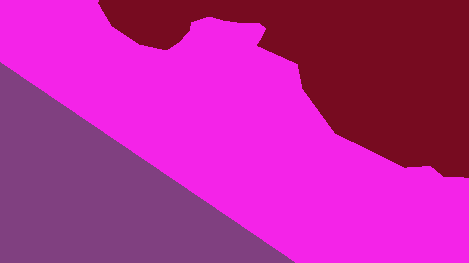}
    \subcaption*{\centering{GT}}
    \label{fig:enter-label11}
    \end{subfigure}
         \hfill
             \begin{subfigure}{0.155\linewidth}
    \includegraphics[width=\linewidth]{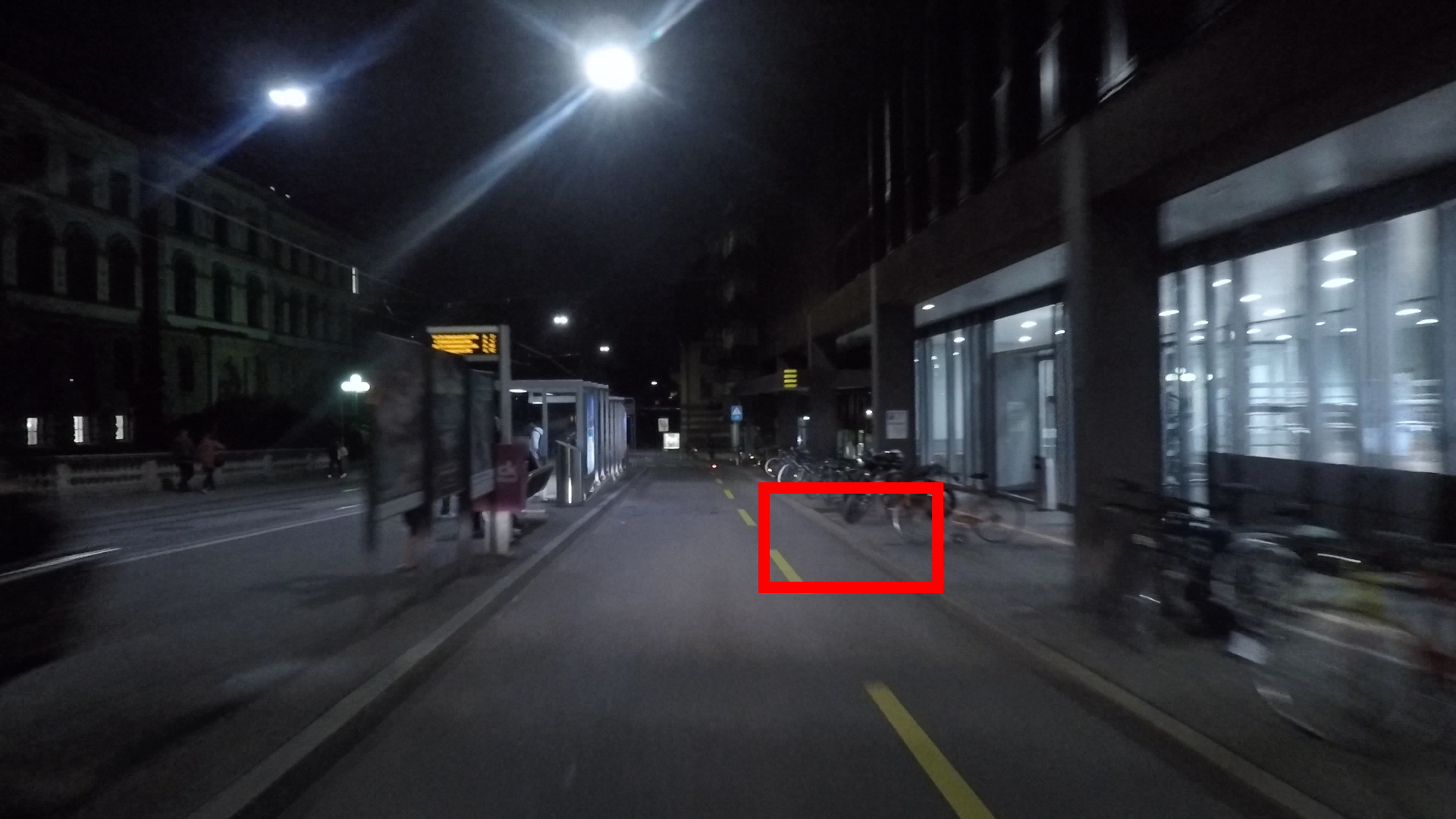}
    \subcaption*{\centering{Input}}
    \label{fig:enter-label1}
    \end{subfigure}
     \hfill
    \begin{subfigure}{0.155\linewidth}
    \includegraphics[width=\linewidth]{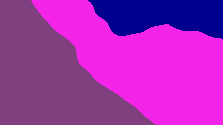}
    \subcaption*{\centering{EnGAN}}
    \label{fig:enter-label2}
    \end{subfigure}
     \hfill
    \begin{subfigure}{0.155\linewidth}
    \includegraphics[width=\linewidth]{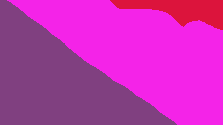}
    \subcaption*{\centering{PairLIE}}
    \label{fig:enter-label3}
    \end{subfigure}
    \hfill
        \begin{subfigure}{0.155\linewidth}
    \includegraphics[width=\linewidth]{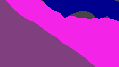}
    \subcaption*{\centering{MBLLEN}}
    \label{fig:enter-label4}
    \end{subfigure}
     \hfill
              \begin{subfigure}{0.155\linewidth}
    \includegraphics[width=\linewidth]{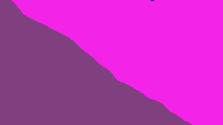}
    \subcaption*{\centering{RetinexNet}}
    \label{fig:enter-label5}
    \end{subfigure}
     \hfill
    \begin{subfigure}{0.155\linewidth}
    \includegraphics[width=\linewidth]{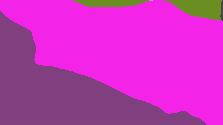}
    \subcaption*{\centering{GLADNet}}
    \label{fig:enter-label6}
    \end{subfigure}
         \hfill
    \begin{subfigure}{0.155\linewidth}
    \includegraphics[width=\linewidth]{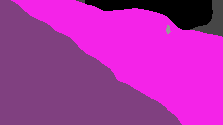}
    \subcaption*{\centering{KinD}}
    \label{fig:enter-label7}
    \end{subfigure}
         \hfill
    \begin{subfigure}{0.155\linewidth}
    \includegraphics[width=\linewidth]{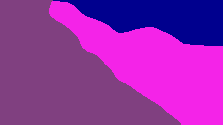}
    \subcaption*{\centering{LLFlow}}
    \label{fig:enter-label8}
    \end{subfigure}
         \hfill
    \begin{subfigure}{0.155\linewidth}
    \includegraphics[width=\linewidth]{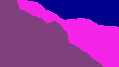}
    \subcaption*{\centering{LLFormer}}
    \label{fig:enter-label9}
    \end{subfigure}
         \hfill
    \begin{subfigure}{0.155\linewidth}
    \includegraphics[width=\linewidth]{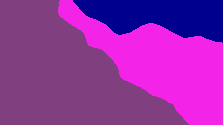}
    \subcaption*{\centering{DRBN}}
    \label{fig:enter-label10}
    \end{subfigure}
         \hfill
    \begin{subfigure}{0.155\linewidth}
    \includegraphics[width=\linewidth]{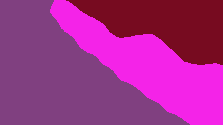}
    \subcaption*{\centering{Ours}}
    \label{fig:enter-label11}
    \end{subfigure}
         \hfill
    \begin{subfigure}{0.155\linewidth}
    \includegraphics[width=\linewidth]{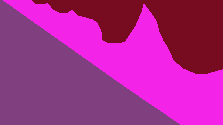}
    \subcaption*{\centering{GT}}
    \label{fig:enter-label11}
    \end{subfigure}
         \hfill
\caption{Visual comparison of segmentation on the ACDC dataset. More results can be found in the Supplementary Materials.}
\label{fig:acdc}
\end{figure}

\noindent\textbf{Evaluaiton on ACDC. }As shown in \Cref{tab:acdc}, LMT-GP has the best mIOU score among all compared methods, with a lead of $3.22\%$  over the second-best method. It is note worthly that only by applying our method or PairLIE can bicycles be detected. We observe that the score of our method exceeds that of PairLIE by 31.8\%. As shown in the zoomed-in regions of \cref{fig:acdc}, LMT-GP yields high-quality segmentation results with reduced artifacts. 

\noindent\textbf{Comparison conclusion. } Experiments on four datasets show that our method has excellent generalization performance. Compared to the state-of-the-art methods, our approach has demonstrated superior performance across various metrics.

\vspace{-0.4cm}
\subsection{Ablation Study}
We conduct ablation studies to evaluate the effectiveness of different loss functions and the influence of PAM. The experiments are performed on the LOL dataset. All models are trained on image patches of size 256 $\times$ 256 with 600 epochs. We provide more ablation results in the Supplementary Materials.

\vspace{-1.3cm}
\begin{table}[H]\scriptsize
\caption{Quantitative results of the loss function evaluation. The best results are in \textcolor{red}{\textbf{red}}.}
\centering
  \begin{tabular}{p{1cm}<{\centering}|p{0.9cm}<{\centering}p{0.9cm}<{\centering}|p{2.2cm}<{\centering}|p{0.8cm}<{\centering}p{0.8cm}<{\centering}|p{1.2cm}<{\centering}p{1.2cm}<{\centering}p{1.2cm}<{\centering}}
    \toprule
    &\multicolumn{2}{c|}{labeled loss}&unlabeled loss&\multirow{2}{*}{$L_{gpr}$}&\multirow{2}{*}{$\hat{L}_{gpr}$}& \multirow{2}{*}{PSNR$\uparrow$}&\multirow{2}{*}{SSIM$\uparrow$} & \multirow{2}{*}{LPIPS$\downarrow$}\\\cline{2-4}
          & $L_{ssim}$ & $L_{p}$ & $L_{un}$ &   &   &   &   &   \\ \hline
        $E_1$ &   & \checkmark & \checkmark &   & \checkmark & 23.87 & 0.901 & \textcolor{red}{\textbf{0.106}} \\ 
        $E_2$ & \checkmark &   & \checkmark &   & \checkmark & 21.66 & 0.896 & 0.177 \\ 
        $E_3$ & \checkmark & \checkmark &   &   & \checkmark & 23.85 & \textcolor{red}{\textbf{0.906}} & 0.116 \\ 
        $E_4$ & \checkmark & \checkmark &   &   &   & 22.00 & 0.897 & 0.120 \\ 
        $E_5$ & \checkmark & \checkmark & \checkmark &  &  & 22.73 & 0.905 & 0.115 \\ 
        $E_6$ & \checkmark & \checkmark & \checkmark & \checkmark &   & 23.55 & 0.905 & 0.114 \\ 
        $E_7$ & \checkmark & \checkmark & \checkmark &   & \checkmark & \textcolor{red}{\textbf{24.12}} & 0.902 & 0.107 \\ \bottomrule
    \end{tabular}
\label{tab:loss}
\vspace{-0.5cm}
\end{table}

\noindent\textbf{A. Evaluation on loss function}\\
\vspace{-0.2cm}

\noindent We evaluate $L_{gpr}$, $\hat{L}_{gpr}$, $L_{ssim}$, $L_{p}$, and $L_{un}$ separately. It is worth noting that the method without $L_{gpr}$ and $\hat{L}_{gpr}$ ($E_5$) implies adopting the basic mean-teacher framework. The quantitative results are shown in \Cref{tab:loss}. Our method ($E_7$) employs an effective loss function strategy that achieves the highest PSNR and SSIM scores. Specifically, by comparing $E_7$ with $E_1$, $E_2$ and $E_3$, we observe that $L_{ssim}$, $L_p$, and $L_{un}$ contribute to the improvement of the PSNR score. Compared to the method without $L_{gpr}$ and $\hat{L}_{gpr}$ ($E_5$), our method ($E_7$) exhibits an improvement of 1.39dB and 0.008 in PSNR and LPIPS scores, respectively. Furthermore, it surpasses the method without ${L}_{gpr}$, $\hat{L}_{gpr}$ and $L_{un}$ ($E_4$) by a significant margin, achieving a PSNR improvement of 2.12dB, an SSIM improvement of 0.005, and an LPIPS improvement of 0.013. $E_6$ and $E_7$ demonstrate that using $\hat{L}_{gpr}$ instead of $L_{gpr}$ leads to a 0.57dB improvement in PSNR score and 0.007 improvement in LPIPS score. 

\begin{figure}[H]
    \begin{subfigure}{0.19\linewidth}
    \includegraphics[width=\linewidth]{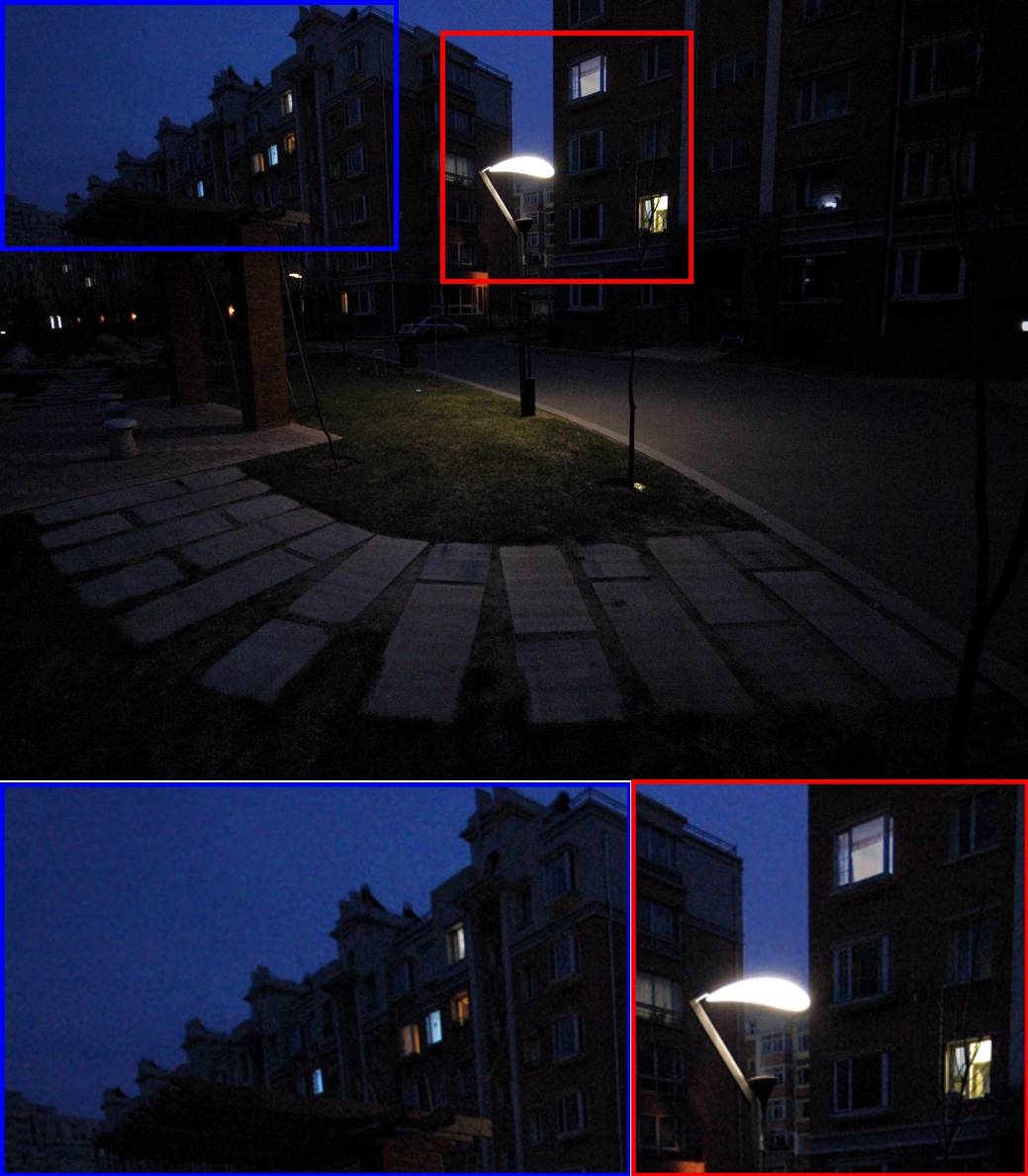}
    \subcaption*{\centering{Input}}
    \label{fig:enter-label7}
    \end{subfigure}
         \hfill
    \begin{subfigure}{0.19\linewidth}
    \includegraphics[width=\linewidth]{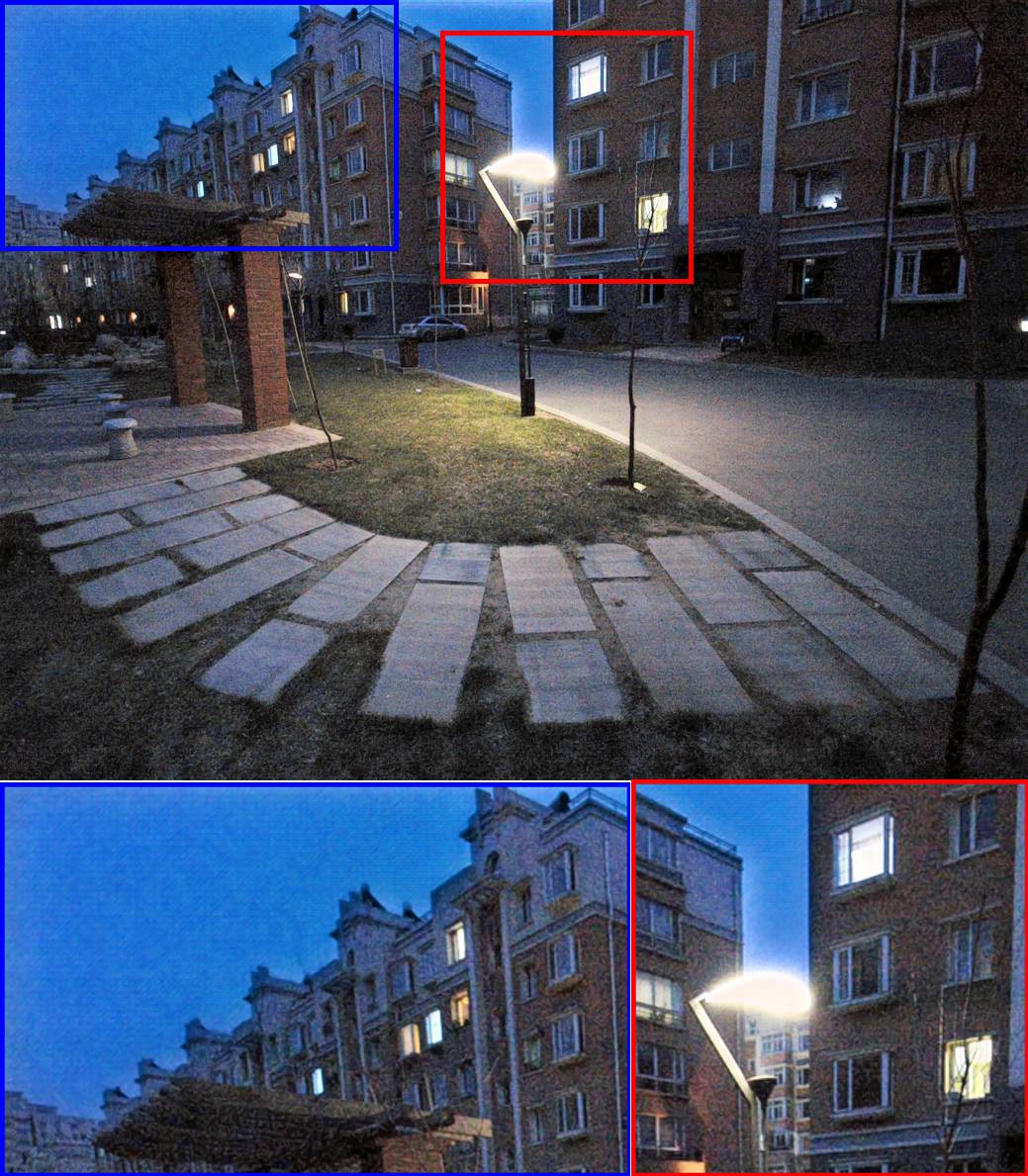}
    \subcaption*{\centering{w/o $L_{ssim}$}}
    \label{fig:enter-label8}
    \end{subfigure}
         \hfill
    \begin{subfigure}{0.19\linewidth}
    \includegraphics[width=\linewidth]{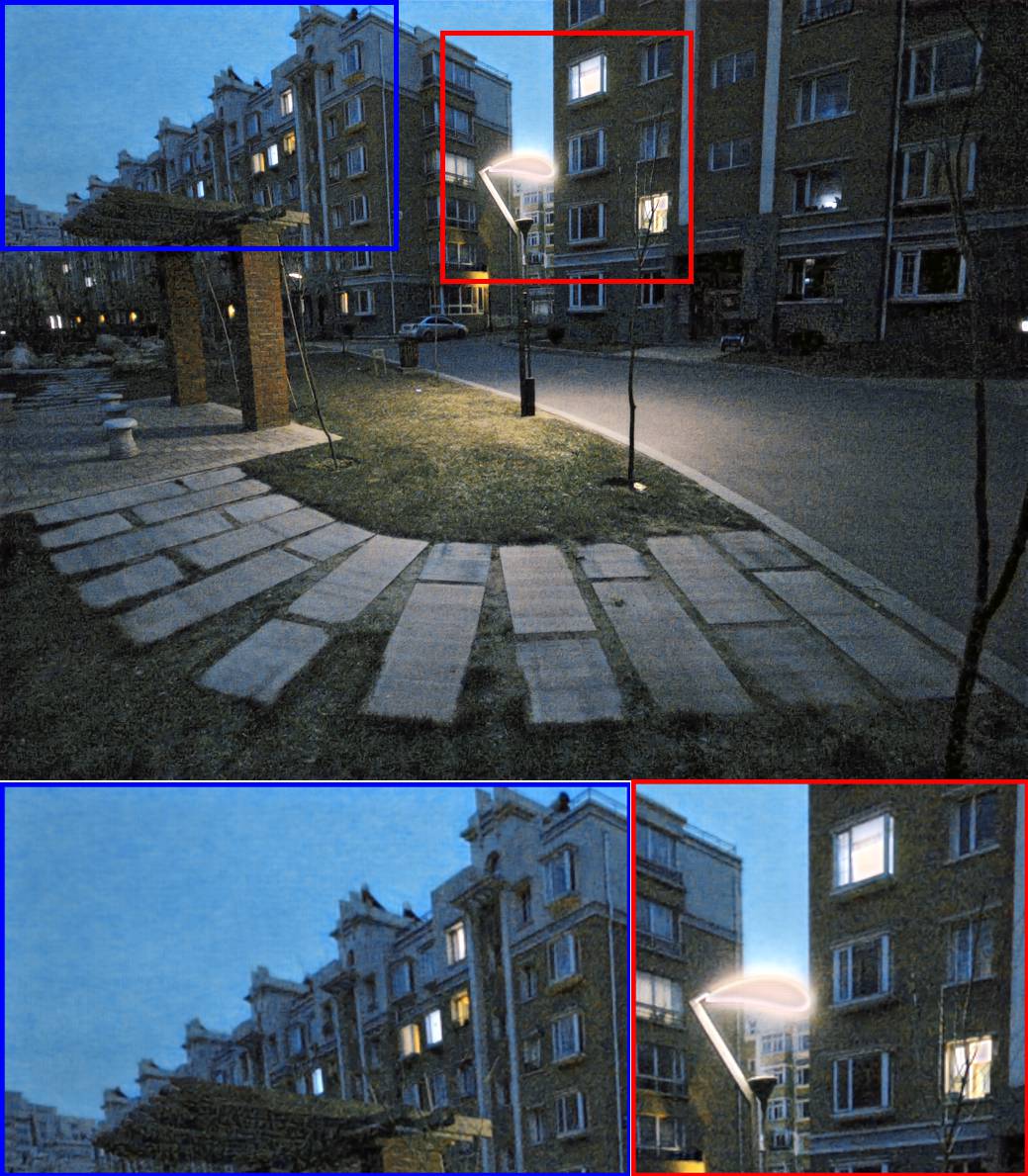}
    \subcaption*{\centering{w/o $L_{p}$}}
    \label{fig:enter-label9}
    \end{subfigure}
         \hfill
    \begin{subfigure}{0.19\linewidth}
    \includegraphics[width=\linewidth]{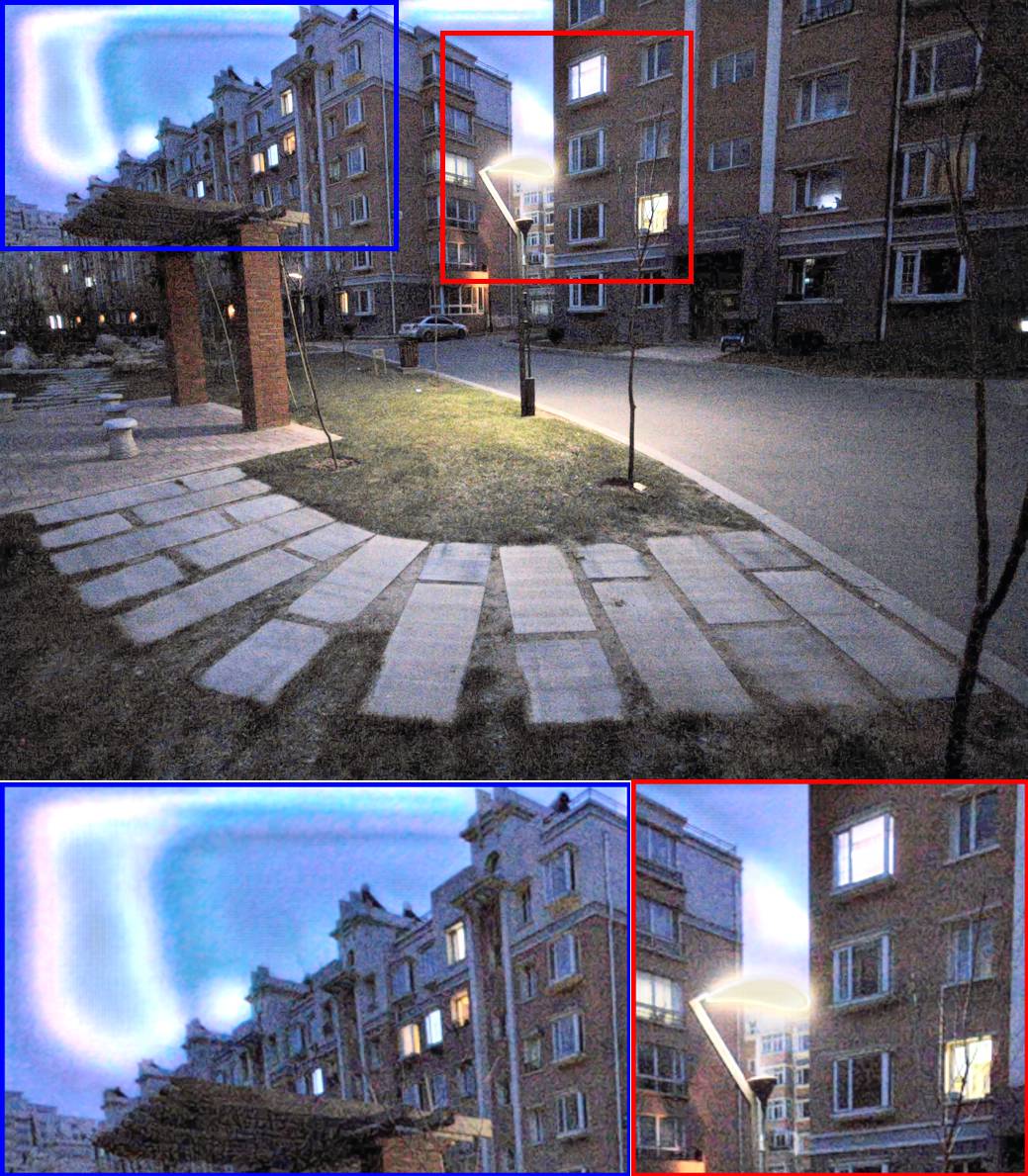}
    \subcaption*{\centering{w/o $L_{gpr}$\& $\hat{L}_{gpr}$}}
    \label{fig:enter-label10}
    \end{subfigure}
         \hfill
    \begin{subfigure}{0.19\linewidth}
    \includegraphics[width=\linewidth]{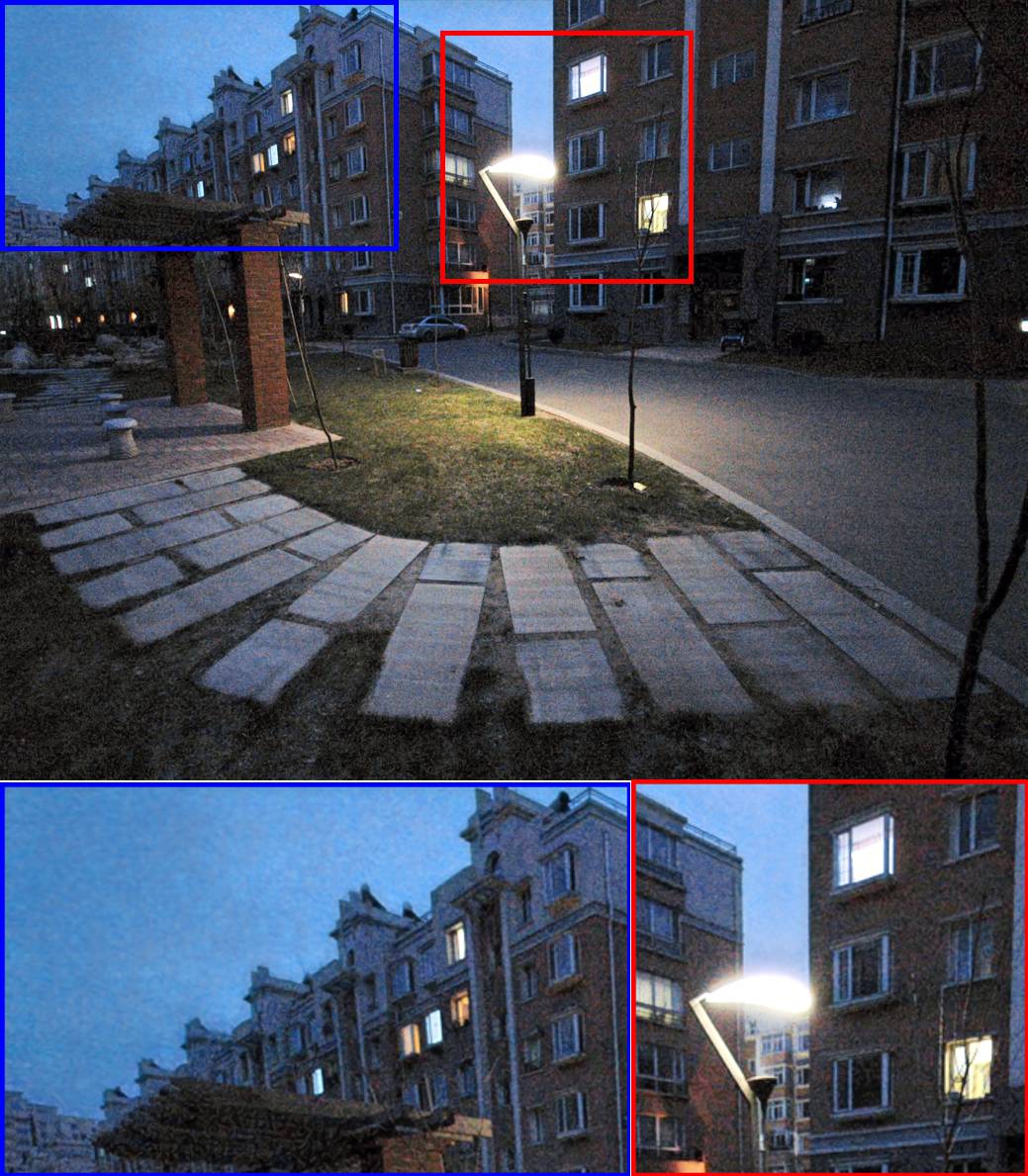}
    \subcaption*{\centering{Ours}}
    \label{fig:enter-label11}
    \end{subfigure}
\caption{Qualitative results of evaluation on different loss function.}
\label{fig:loss}
\end{figure}

The qualitative results are shown in \cref{fig:loss}. The method without $L_{ssim}$ introduces noise and the method without $L_{p} $ results in artifacts. Compared to the method without $L_{gpr}$ and $\hat{L}_{gpr}$, our method achieves better results in recovering
\vspace{-1.25cm}
\begin{table}[H]\scriptsize
\caption{Quantitative results of PAM and influence of IQA metrics. The best results are in \textcolor{red}{\textbf{red}}.}
\centering
 \begin{tabular}{p{1.2cm}<{\centering}|p{1.0cm}<{\centering}p{1.0cm}<{\centering}p{1.0cm}<{\centering}|p{1.5cm}<{\centering}p{1.5cm}<{\centering}p{1.5cm}<{\centering}}
    \toprule
          & \multicolumn{3}{c|}{IQA metrics}   & \multirow{2}{*}{PSNR$\uparrow$} & \multirow{2}{*}{SSIM$\uparrow$} & \multirow{2}{*}{LPIPS$\downarrow$} \\ \cline{2-4}
          & PSNR & SSIM & LPIPS &   &   \\ \hline
        w/o PAM &   &   &   & 21.54 & 0.897 & 0.127\\ \hline
        \multirow{3}{*}{w/ PAM}  &  \checkmark &   &   &  \textbf{ \textbf{ \textcolor{red}{24.12}}} & 0.902 & \textcolor{red}{\textbf{0.107}}\\ 
          &   &  \checkmark &   & 22.65 &  \textbf{ \textbf{ \textcolor{red}{0.905}}} &0.116\\ 
          &   &   &  \checkmark & 23.96 & 0.904 &0.113\\ \bottomrule
    \end{tabular}
\label{tab:metrics}
 \vspace{-0.4cm}
\end{table}

\noindent low-level features such as color and texture.

\vspace{0.3cm}
\noindent\textbf{B. Evaluation on PAM}\vspace{0.2cm}
\\
We conduct experiments to evaluate the impact of PAM and different image quality assessment (IQA) metrics used to selected the latent vectors for PAM. The results are shown in \Cref{tab:metrics}. Compared to the method without PAM,  PAM improves the PSNR, SSIM, and LPIPS scores of the enhanced images. We also observe that PAM achieves the best performance when using PSNR metrics to pinpoint the latent vectors.

\section{Conclusion}
In this paper, we propose a novel semi-supervised method to improve the visual quality of low-light images. We introduce a latent mean-teacher framework to effectively incorporate latent vectors into network learning. Meanwhile, we design a mean-teacher assisted GP learning strategy to address the domain issue. This strategy uses labeled data to represent the latent vectors of unlabeled data, establishing a connection between the latent vectors of unlabeled and labeled data. We also design an assisted GPR loss to learn the distribution of unlabeled data. Furthermore, we design a PAM to guarantee the reliability of the network learning. Compared with existing methods, Our method effectively captures the illumination of low-light and normal-light images. Additionally, we conduct extensive experiments on high-level visual tasks. From a task-oriented perspective, our method achieves remarkable performance in low-level vision and shows strong generalizability. We believe that our method can further advance the development of semi-supervised LLIE research in the future.

\clearpage  

\section*{Acknowledgements}
This work is supported by grants of the National Natural Science Foundation of China (62372153, 62276242), the Anhui Provincial Natural Science Foundation (2308085MF216), the National Aviation Science Foundation (2022Z071078001), and the Dreams Foundation of Jianghuai Advance Technology Center (2023-ZM01Z001).

%
%
\bibliographystyle{splncs04}
\bibliography{main}
\end{document}